\documentclass[11pt]{article}

\usepackage{microtype}
\usepackage{graphicx}
\usepackage{subcaption}
\usepackage{booktabs}
\usepackage{microsoft-tech-report}
\usepackage{hyperref}
\usepackage{url}
\usepackage[utf8]{inputenc}
\usepackage[T1]{fontenc}
\usepackage{amsmath}
\usepackage{amssymb}
\usepackage{amsfonts}
\usepackage{mathtools}
\usepackage{amsthm}
\usepackage{bm}
\usepackage{nicefrac}
\usepackage{enumitem}
\usepackage{multirow}
\usepackage[capitalize,noabbrev]{cleveref}
\usepackage{xspace}
\usepackage{pifont}
\usepackage{wrapfig}
\usepackage{algorithm}
\usepackage{algpseudocode}
\usepackage{placeins}

\definecolor{deltapos}{HTML}{0E8A4A}
\definecolor{deltaneg}{HTML}{B91C1C}

\definecolor{avglowbg}{HTML}{FDE8E8}
\newcommand{\avglow}[1]{\cellcolor{avglowbg}#1}
\newcommand{\avgbest}[1]{\textcolor{msftblue}{\textbf{#1}}}
\newcommand{\std}[1]{{\fontsize{6pt}{6pt}\selectfont\textcolor{gray}{$\pm$#1}}}

\newtcolorbox{findingbox}[1][]{
  enhanced,
  colback=msftcard,
  colframe=msftblue,
  boxrule=0.6pt,
  arc=2pt,
  left=8pt, right=8pt, top=9pt, bottom=6pt,
  before skip=14pt, after skip=14pt,
  fontupper=\normalsize\color{msftdark},
  attach boxed title to top left={xshift=8pt, yshift=-\tcboxedtitleheight/2},
  boxed title style={
    colback=msftcard,
    colframe=msftblue,
    boxrule=0.6pt,
    arc=3pt,
    left=6pt, right=6pt, top=2pt, bottom=2pt,
  },
  title={\msftsans\bfseries\normalsize\color{msftblue}Takeaway},
}
\newcommand{\findinglead}[1]{{\msftsans\bfseries\color{msftblue}Finding\ifx\\#1\\\else~(#1)\fi.}\ }

\newtcolorbox{promptbox}[1][]{
  enhanced,
  breakable,
  colback=msftcard,
  colframe=msftblue,
  boxrule=0.5pt,
  arc=2pt,
  left=6pt, right=6pt, top=4pt, bottom=4pt,
  before skip=8pt, after skip=8pt,
  fonttitle=\msftsans\bfseries\color{white},
  colbacktitle=msftblue,
  title={#1},
}

\hypersetup{
  colorlinks=true,
  linkcolor=msftblue,
  citecolor=msftblue,
  urlcolor=msftblue,
  pdftitle={AgentStream}
}

\begin{document}
\thispagestyle{empty}

\noindent
\begin{minipage}[c]{0.5\linewidth}
\raggedright
\raisebox{-0.5\height}{\msftbrandmark}
\end{minipage}%
\begin{minipage}[c]{0.49\linewidth}
\raggedleft
{\msftdatefont\small\color{msftgray}July 2026}
\end{minipage}\par
\vspace{0.35em}
\noindent{\color{msftline}\rule{\linewidth}{0.8pt}\par}

\vspace{1.0em}
\begin{center}
{{\msfttitlefont\fontsize{21}{25}\selectfont\color{msftdark}
AgentStream: How Well Do Self-Evolving LLM Agents Perform Under Streaming Tasks?
}}
\vspace{1.25em}

{\normalsize\rmfamily\color{msftdark}
Dong Yan$^{1,2,3,\star}$ \hspace{0.6em}
Jian Liang$^{1,3,\ddagger}$ \hspace{0.6em}
Dapeng Hu$^{2,\ddagger}$ \hspace{0.6em}
Ran He$^{1,3}$ \\[-0.1em]
Nicholas Jing Yuan$^{2}$ \hspace{0.6em}
Qi Zhang$^{2}$ \hspace{0.6em}
Tieniu Tan$^{1,3,4}$ \hspace{0.6em}
}
\vspace{0.22cm}

{\footnotesize\rmfamily\color{msftgray}
$^{1}$ School of Artificial Intelligence, University of Chinese Academy of Sciences \quad
$^{2}$ Microsoft \quad
\\
$^{3}$ Institute of Automation, Chinese Academy of Sciences \quad
$^{4}$ Nanjing University \par
}
\end{center}

\vspace{0.45em}
\begin{msfttitlebox}
\setlength{\parindent}{0cm}
\setlength{\parskip}{0.14cm}
\raggedright
\nohyphens

\begin{abstract}
Large language model (LLM) agents can self-evolve by continually improving from their own accumulated experience.
However, existing studies predominantly adopt independent evaluation.
Consequently, the behavior of self-evolving agents in realistic streaming settings, where agents adapt to diverse and complex task streams, remains poorly understood.
To address this gap, we introduce AgentStream, a unified framework that evaluates self-evolving agents spanning diverse evolution components by organizing agentic benchmarks into a configurable task stream.
Specifically, we instantiate the \texttt{Isolated}, \texttt{Sequential}, and \texttt{Interleaved} streaming scenarios at test time, which progressively vary the scope and domain composition of the stream.
Over these scenarios, we combinatorially evaluate five representative self-evolving methods across three frontier foundation models, disentangling how model capability, method architecture, and streaming scenario jointly shape self-evolution.
Our results show that self-evolution reliability varies across streaming scenarios, the benefit of self-evolution is gated by model capability and is non-monotonic in model strength, and no single method dominates across models and scenarios.
These findings offer concrete guidance for selecting self-evolving methods across models and streaming scenarios.
Overall, we advocate that self-evolving agents should be evaluated under realistic task streams rather than isolated single-task settings.
\end{abstract}
\vspace{0.14cm}
{\setlength{\parskip}{0.06cm}\small
{\msftmetalabel{Code}\href{https://github.com/microsoft/Sico/tree/main/labs/AgentStream}{https://github.com/microsoft/Sico/labs/AgentStream}\par}
}
\end{msfttitlebox}

\renewcommand{\thefootnote}{}%
\makeatletter
\long\def\@makefntext#1{\noindent#1}%
\footnotetext{$^{\star}$ Work done during an internship at Microsoft.\par
$^{\ddagger}$ Corresponding authors: \href{mailto:liangjian92@gmail.com}{liangjian92@gmail.com}, \href{mailto:dapenghu@microsoft.com}{dapenghu@microsoft.com}.}%
\makeatother
\renewcommand{\thefootnote}{\arabic{footnote}}%


\section{Introduction}
Large language model (LLM) agents are shifting from static, deploy-once systems toward adaptive architectures that continuously learn from their own accumulated experience during deployment, a paradigm broadly termed self-evolving~\citep{gao2026a, lin2026position, fang2025comprehensive,team2026agentic}.
Depending on which component of the agent is updated, recent methods evolve the prompt context, structured memory, reusable skill libraries, or an integrated harness~\citep{zhang2026ace,lin2026harness,xu2025amem, zhang2026memrl,yang2026autoskill, ouyang2026reasoningbank, zhou2026mementoskills,lin2026ahe}.
Such self-evolution is expected to produce increasingly capable agents over time, making it essential to understand how agent capabilities evolve across the diverse task streams encountered in realistic deployment.

\begin{figure*}[t]
\vspace{-1.5\baselineskip}
    \centering
    \includegraphics[width=1.0\textwidth]{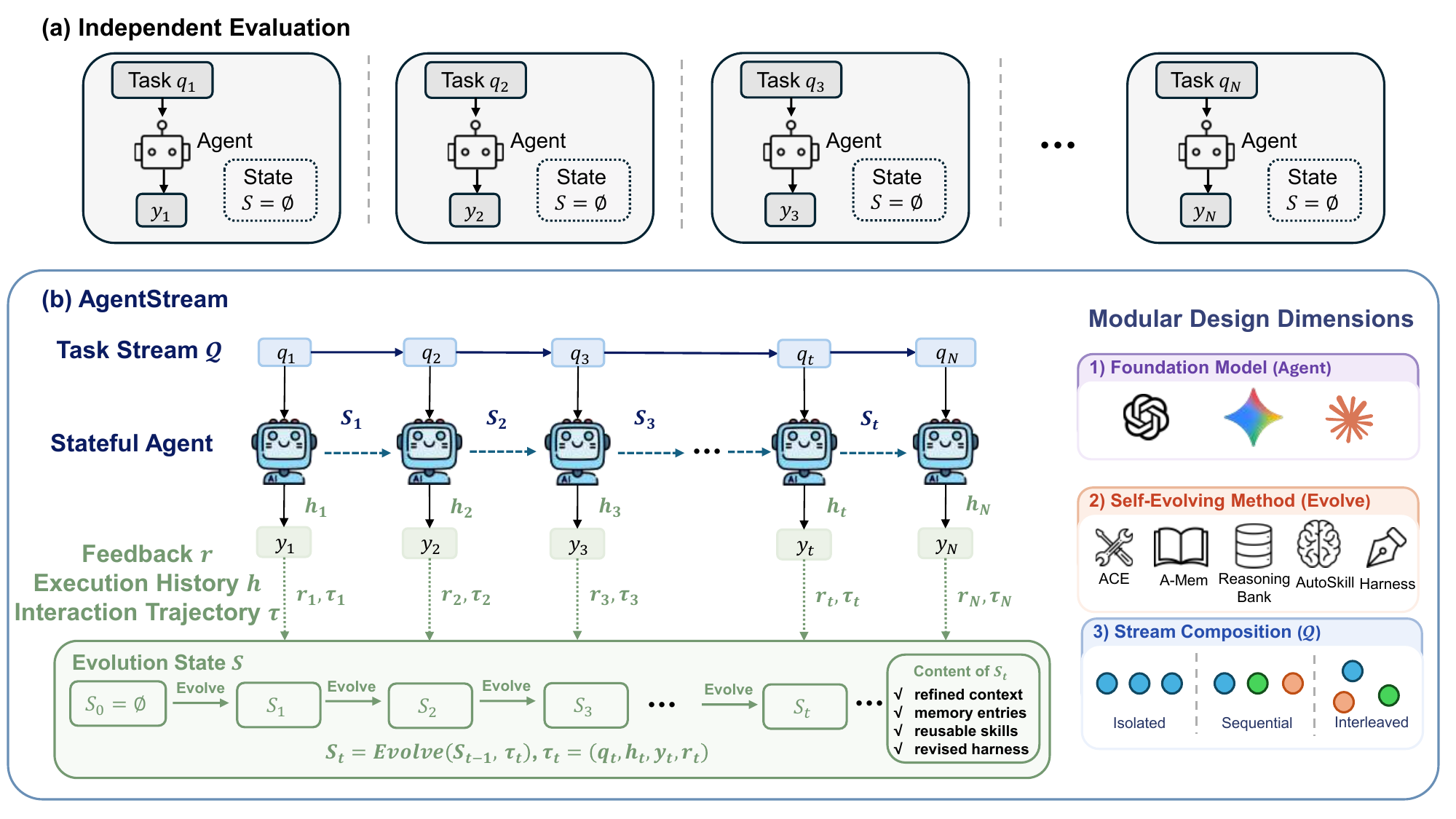}
    \caption{Independent evaluation vs. AgentStream framework. (a) The prevailing paradigm solves each task in isolation without cross-instance experience accumulation. (b) AgentStream: a stateful agent evolves through self-generated feedback over a configurable task stream, with foundation model, self-evolving method, and stream composition as modular design dimensions.}
    \label{fig:compare}
\end{figure*}

However, as illustrated in \cref{fig:compare}, existing agentic benchmarks and self-evolving studies predominantly adopt independent evaluation, where each task is solved in isolation and performance is aggregated without any cross-task state~\citep{appworld, bfcl, jimenez2024swe, hle, barres2025tau, zhang2026ace,xu2025amem}.
While a few studies move toward streaming evaluation~\citep{wu2024streambench, wei2026evomemory,ouyang2026reasoningbank}, they stream each benchmark independently under a single evolution component, leaving how different evolution components transfer cross-domain experience systematically unexamined.
Consequently, it remains unclear whether the improvements reported for self-evolving agents persist once they are deployed in realistic streaming settings, where tasks may span diverse domains and arrive without clear domain boundaries or supervision.
Answering this question requires an evaluation that considers the foundation model, the self-evolving method, and the structure of the task stream jointly rather than any one of them in isolation, which the prevailing independent evaluation is inherently unable to provide.

To this end, we propose AgentStream, a unified streaming evaluation framework that organizes tasks from multiple agentic benchmarks into a configurable stream, ranging from within-domain to cross-domain composition, and evaluates self-evolving methods whose evolution components span context, memory, skill, and integrated harness.
AgentStream performs a combinatorial evaluation across models, self-evolving methods, and streaming scenarios, enabling us to decouple the contributions of model capability and method architecture under different stream structures.
Concretely, we instantiate three test-time streaming scenarios that progressively vary in scope and domain composition: \texttt{Isolated}, where the agent evolves independently within each benchmark; \texttt{Sequential}, where the agent processes the benchmarks in a fixed order, transferring its evolution state across benchmarks; and \texttt{Interleaved}, where tasks from all benchmarks are shuffled into one unified stream.
Across these scenarios, we evaluate five representative self-evolving methods, ACE~\citep{zhang2026ace}, A-Mem~\citep{xu2025amem}, ReasoningBank~\citep{ouyang2026reasoningbank}, AutoSkill~\citep{yang2026autoskill}, and Harness~\citep{lin2026harness, lin2026ahe}, across three frontier foundation models, GPT-5.4~\citep{singh2025openai}, Gemini 3.1 Pro~\citep{google2026gemini31pro}, and Claude Opus 4.7~\citep{anthropic2026claudeopus47}, over six agentic benchmarks covering diverse capabilities, including AppWorld~\citep{appworld}, BFCL~\citep{bfcl}, BrowseComp-Plus~\citep{browsecompplus}, HLE~\citep{hle}, SWE-bench Verified~\citep{jimenez2024swe}, and Tau2~\citep{barres2025tau}.

Our study reveals three findings.
\textbf{First}, self-evolution is not uniformly beneficial, and its reliability varies with the streaming scenario. \texttt{Isolated} is the most reliable, while \texttt{Interleaved} generally outperforms \texttt{Sequential} despite its more heavily mixed stream.
\textbf{Second}, the benefit of self-evolution is gated by model capability. The weakest model exhibits negative evolution gains, and the benefit is non-monotonic in model strength, with a mid-capability model gaining more than a stronger one.
\textbf{Third}, no single method dominates. Context-integrated methods favor \texttt{Isolated} while retrieval-based methods favor \texttt{Interleaved}, and the optimal method varies across models rather than transferring between them.

Overall, this work contributes AgentStream, the first framework that unifies agentic benchmarks into a configurable streaming evaluation and systematically assesses self-evolution along the streaming scenario, method, and model dimensions. Through a combinatorial analysis over these dimensions, we investigate how streaming scenario, model capability, and method architecture shape shape the performance of self-evolution.  We further distill these observations into actionable guidance: applying self-evolution to sufficiently capable models, favoring context-integrated methods under within-domain streams and retrieval-based methods under cross-domain streams, and selecting the method per model rather than assuming a universal choice. We hope AgentStream encourages future research to evaluate self-evolving agents beyond isolated single-task settings and toward realistic task streams.

\section{Related Work}
\subsection{Learning with Streaming Data}
Learning from streaming data mainly involves test-time learning and continual learning.
Test-time learning adapts a model to each incoming instance or distribution shift on the fly~\citep{sun2020test,liang2020we}, whereas continual learning targets sequential task streams while resisting catastrophic forgetting~\citep{kirkpatrick2017overcoming, lopez2017gradient}.
With the advent of LLMs and agents, test-time learning proceeds at two levels~\citep{snell2024scaling, hardt2024nearestneighbors}.
Parameter adaptation methods operate within individual test instances, either by directly updating model weights through data selection and in-place learning~\citep{hubotter2024efficiently,  akyurek2025surprising, feng2026inplace, chen2026testtimeadaptation, hu2025ttl}, or by applying reinforcement learning to iteratively refine model behavior~\citep{zuo2025ttrl, hubotter2025learningjob, yuksekgonul2026learningdiscover, hu2026collaborative, yan2026scrl, he2026ttsr, yang2026ttcs}.
Cross-instance accumulation methods instead build reusable experience that transfers across test tasks, including evolving libraries~\citep{xu2026evolvinglibrary, wei2026evomemory}, temporary skills~\citep{wang2026skillsfly, wang2026tarse}, cached plan templates~\citep{zhang2026plancaching}, and consolidated memory systems~\citep{cheng2026tame, acikgoz2025selfimproving}.
In parallel, continual learning for LLM agents has been explored through gradient-free inference-time updates~\citep{li2026just, rajesh2026panini}, parameter-level methods that mitigate forgetting~\citep{wu2026agent, lu2026mssr}, and experience-centric approaches that reuse accumulated knowledge across tasks~\citep{xiong2026learning, jiang2026xskill, ye2026online, yang2026adaptive}.

\subsection{Self-Evolving Agents}
Self-evolving agents can autonomously improve their capabilities by learning from their own experience~\citep{gao2026a, lin2026position}. 
Existing approaches primarily differ in the component of the agent system that undergoes evolution.
Memory-based methods focus on the accumulation, retrieval, and reorganization of historical experience, enabling agents to selectively retain useful knowledge while pruning irrelevant information over successive tasks~\citep{zhang2026memrl, zhang2026memskill, xu2025amem, chhikara2025mem0, zhang2026ace}. 
Skill-based methods operate at a higher level of abstraction, discovering and composing reusable capability modules into growing libraries that compound across diverse task domains~\citep{xia2026skillrl, yang2026autoskill, ouyang2026skillos, alzubi2026evoskill, zhou2026mementoskills, yan2026openskill, yang2026skillopt}. 
Harness-level methods evolve the scaffolding infrastructure surrounding the base model, including system prompts, tool configurations, memory modules, and skill libraries~\citep{lee2026metaharness, guo2026evoconfig, lin2026harness, lin2026ahe, zhang2026selaur, xu2026bes, acikgoz2026toolr0, sheng2026rlcer}. 
Beyond individual agents, architecture-level methods evolve the multi-agent topology itself, dynamically reconfiguring agent roles, communication structures, and coordination protocols~\citep{wang2026metagen, hu2024evomac, liu2026sew}. 
Mechanistically, these methods are driven by reinforcement learning~\citep{zhang2026memrl, xia2026skillrl, zhang2026selaur}, evolutionary algorithms~\citep{yang2026evotool,  novikov2025alphaevolve, xu2026bes}, trajectory distillation~\citep{yang2026autoskill, he2026evotest, cai2026ell}, or gradient-analogy optimization that treats textual feedback as differentiable signals~\citep{yuksekgonul2025textgrad, hu2024evomac,yang2026skillopt}.
However, existing work evaluates on isolated benchmarks without systematically comparing how different self-evolving methods and models behave under a controlled setting, while recent analyses further reveal that self-evolution can degrade or fail to transfer across domains~\citep{shao2026misevolve,gao2026a, fang2025comprehensive}.
Our work provides a unified streaming evaluation framework, analyzing the roles of self-evolving methods and models under different stream structures.

\subsection{Agentic Benchmarks}

Agentic benchmarks have been developed to evaluate LLM agents across diverse and complex environments.
Interactive web and application benchmarks require agents to complete long-horizon tasks by navigating stateful interfaces and executing actions that alter the environment state~\citep{yao2022webshop, zhou2024webarena, he2024webvoyager, appworld, xie2024osworld}.
Tool-use benchmarks assess structured function calling, measuring whether agents can select, compose, and invoke external APIs with correct arguments~\citep{bfcl, bandi2026mcpatlas,barres2025tau,li2025tool}.
Software engineering benchmarks evaluate agents on repository-level tasks such as resolving real-world code issues under executable test suites~\citep{jimenez2024swe, tbench2026terminal, wang2026cybergym}.
Skill-oriented benchmarks measure how well agents acquire and reuse modular capabilities across heterogeneous tasks~\citep{li2026skillsbench, zhang2026skillflow}.
Knowledge-intensive reasoning benchmarks stress deep retrieval and expert-level problem solving over frontier knowledge~\citep{hle, mialon2023gaia, wei2025browsecomp}.
Economically grounded benchmarks further evaluate agents on real-world professional tasks of high practical value~\citep{patwardhan2025gdpval,bigeard2025finance}.
However, these benchmarks predominantly evaluate agents on each task in isolation, and although a few adopt streaming evaluation tailored for self-evolving agents~\citep{wu2024streambench, wei2026evomemory,wang2025agent}, they remain limited to single-benchmark streams and evaluate only one evolving component.
In contrast, our framework unifies agentic benchmarks into more complex streaming scenarios covering both within-domain and cross-domain adaptation, and evaluates diverse evolving components including prompt, memory, skill, and harness.

\begin{figure*}[t]
    \centering
    \includegraphics[width=1.0\textwidth]{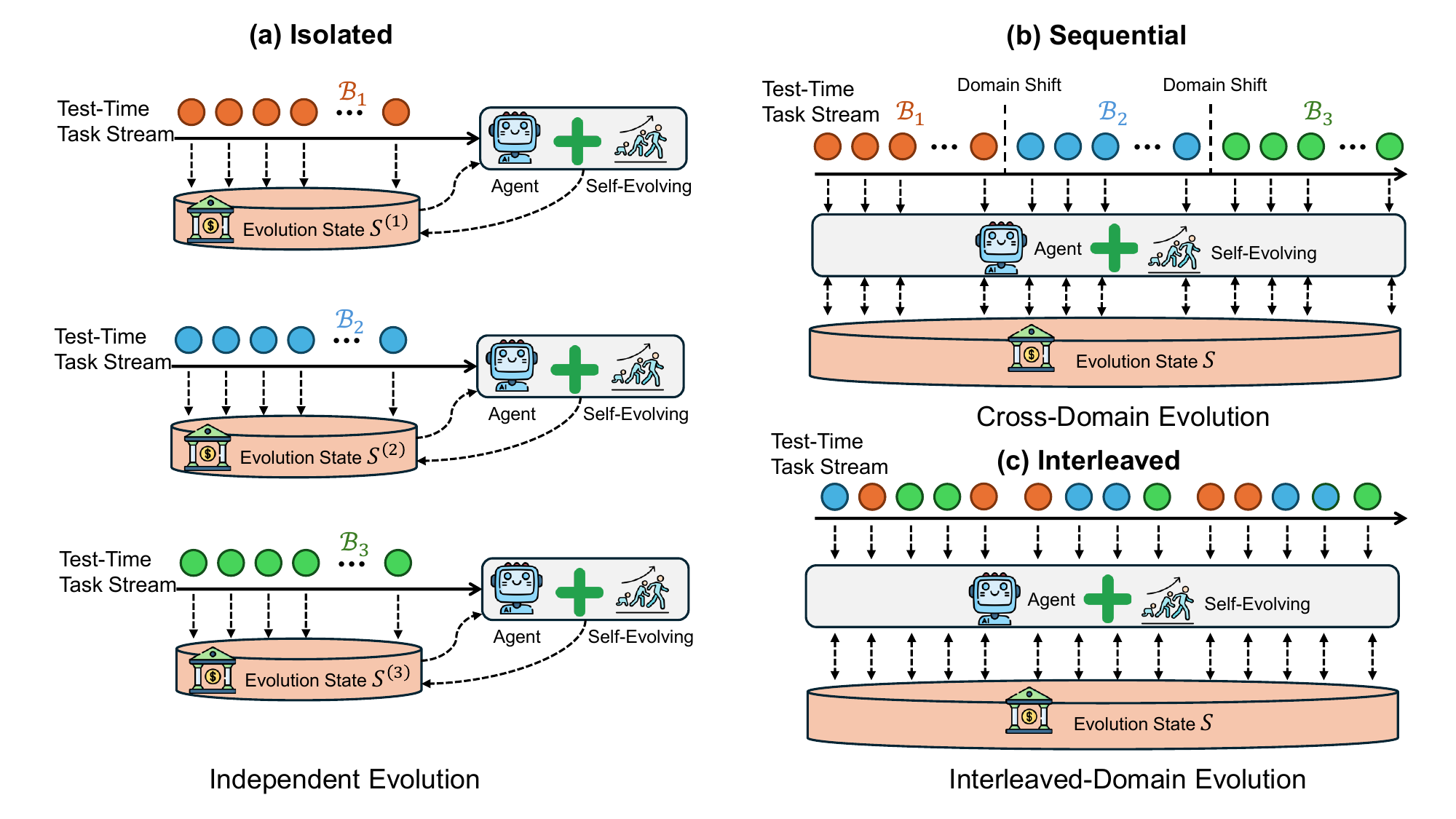}  
    \caption{Three representative streaming scenarios instantiated in AgentStream. (a) \texttt{Isolated}: each benchmark maintains a separate evolution state. (b) \texttt{Sequential}: a single evolution state carries across ordered benchmark boundaries. (c) \texttt{Interleaved}: tasks from all benchmarks are shuffled into one unified stream with a shared evolution state.}
    \label{fig:overview}
\end{figure*}

\section{The AgentStream Framework}
To study how LLM agents improve through experience accumulation over realistic task streams, we design AgentStream, an evaluation framework centered on self-evolution, in which an agent processes a stream of tasks and, after each attempt, distills the interaction trajectory into persistent experience (e.g., refined context, memory entries, reusable skills, or revised harness).
Building on this framework, we instantiate the problem setup and three streaming scenarios that systematically vary the scope and composition of the task stream, as illustrated in \cref{fig:overview}.

\subsection{Problem Setup}
We consider a test-time learning setting~\citep{ouyang2026reasoningbank,wu2024streambench} in which a task stream $\mathcal{Q} = \{q_1, q_2, \ldots, q_N\}$ is presented to the agent, with each task revealed only after the preceding one is completed. 
The agent is parameterized by a foundation model $\mathcal{M}$ and equipped with a self-evolving method that maintains an evolution state $S$. 
This state aggregates the experience distilled from the first $t$ interactions, initialized as $S_0 = \varnothing$. 
For each task $q_t$, the agent performs a multi-step interaction conditioned on its current evolution state, generating an execution history $h_t = \{(a_{t,i}, o_{t,i})\}_{i=1}^{L_t}$ of action-observation pairs over $L_t$ steps and a final solution $y_t = \mathcal{M}(q_t, h_t \mid S_{t-1})$. The agent subsequently updates $S_t = \textsc{Evolve}(S_{t-1},\, \tau_t)$ by reflecting on the interaction trajectory $\tau_t = (q_t, h_t, y_t, r_t)$, where $r_t$ denotes the self-generated feedback. 
Notably, no ground-truth labels are accessible at test time, and the evolution relies entirely on feedback intrinsic to the interaction, such as execution outcomes and reflective self-evaluation.
We quantify the benefit of self-evolution through the evolution gain:
\begin{equation}
  \Delta = \mathrm{Perf}(\mathcal{M},\, \mathcal{Q},\, S) \;-\; \mathrm{Perf}(\mathcal{M},\, \mathcal{Q},\, \varnothing),
\end{equation}
where $\mathrm{Perf}(\mathcal{M}, \mathcal{Q}, \varnothing)$ denotes the baseline in which the same model solves each task with $S_t = \varnothing$ for all $t$. 
A positive $\Delta$ indicates net improvement from self-evolution, while a negative $\Delta$ signals that accumulated experience introduces interference.

\subsection{Streaming Scenarios}
\label{sec:three_level}
Given a set of $K$ benchmarks $\{\mathcal{B}_1, \ldots, \mathcal{B}_K\}$, each contributing a task subset $\mathcal{Q}^{(k)}$, we instantiate three evaluation settings that systematically vary how the task stream $\mathcal{Q}$ is composed and how the evolution state $S_t$ is scoped across benchmarks.

\paragraph{\texttt{Isolated}.}
Each benchmark $\mathcal{B}_k$ is assigned an independent agent instance with its own evolution state $S_t^{(k)}$. 
The task stream for each agent instance is simply $\mathcal{Q}^{(k)}$, and no experience transfers across benchmarks. 
This setting isolates intra-domain evolution and measures how effectively a self-evolving method accumulates useful experience within a single task domain.

\paragraph{\texttt{Sequential}.}
A single agent processes all benchmarks in a fixed order $\mathcal{B}_1 \rightarrow \mathcal{B}_2 \rightarrow \cdots \rightarrow \mathcal{B}_K$, with its evolution state $S_t$ retained across benchmark boundaries. 
The resulting task stream is the concatenation $\mathcal{Q} = \mathcal{Q}^{(1)} \oplus \mathcal{Q}^{(2)} \oplus \cdots \oplus \mathcal{Q}^{(K)}$. 
This setting evaluates whether experience acquired in earlier domains facilitates or interferes with performance on later domains, testing forward transfer in a sequential curriculum.

\paragraph{\texttt{Interleaved}.}
A single agent receives a unified stream $\mathcal{Q} = \mathrm{shuffle}(\mathcal{Q}^{(1)} \cup \cdots \cup \mathcal{Q}^{(K)})$ in which tasks from all benchmarks are interleaved in randomized order. 
The agent maintains one shared evolution state $S_t$ that is updated by tasks from all benchmarks indiscriminately. This setting tests the agent's ability to retrieve domain-relevant experience while suppressing cross-domain interference under maximal task diversity.

Rather than being ordered by expected difficulty, the three streaming scenarios are designed to decouple distinct aspects of self-evolution, each capturing a challenge that self-evolving agents may encounter in deployment.
\texttt{Isolated} removes cross-domain effects entirely, providing a controlled measurement of within-domain learning.
\texttt{Sequential} introduces ordered domain shifts and tests whether accumulated experience enables forward transfer. 
\texttt{Interleaved} mixes tasks from all domains within a single stream, requiring the agent to retrieve relevant experience and suppress irrelevant interference without explicit domain boundaries.
Crossing these three streaming scenarios with the self-evolving methods and frontier foundation models enables a controlled analysis of how each factor contributes to self-evolution.

\section{Experimental Settings}

\paragraph{Models.}
We evaluate three frontier foundation models spanning different families and scales: GPT-5.4-medium~\citep{singh2025openai}, Gemini 3.1 Pro-medium~\citep{google2026gemini31pro}, and Claude Opus 4.7-high~\citep{anthropic2026claudeopus47}. 
This diversity allows us to analyze how model capacity influences test-time evolution across self-evolving methods.
\paragraph{Tasks.}
Our evaluation suite comprises six diverse benchmarks covering a broad spectrum of agentic capabilities:
\begin{itemize}[leftmargin=*, nosep]
\item \textbf{AppWorld}~\citep{appworld}: interactive coding tasks requiring multi-app workflow execution with API understanding and dynamic environment interaction.
\item \textbf{BFCL}~\citep{bfcl}: multi-step function calling evaluation across diverse domains, testing context-dependent tool use with missing parameters and long-context scenarios.
\item \textbf{BrowseComp-Plus}~\citep{browsecompplus}: deep-research tasks requiring iterative web retrieval and complex information synthesis over a controlled document corpus.
\item \textbf{HLE}~\citep{hle}: expert-level academic reasoning across dozens of disciplines, designed to challenge frontier models on questions resistant to memorization.
\item \textbf{SWE-bench Verified}~\citep{jimenez2024swe}: real-world software engineering tasks requiring codebase understanding, fault localization, and patch generation.
\item \textbf{Tau2}~\citep{barres2025tau}: conversational agent tasks in dual-control environments where both agent and user take actions in a shared system, testing coordination and communication.
\end{itemize}

\paragraph{Self-Evolving Methods.}
We select five representative methods that span context, memory, skill, and integrated harness evolution, collectively representing the principal evolution components:
(1) \textbf{ACE}~\citep{zhang2026ace}: evolves agent context through modular generation, reflection, and curation of prompts, accumulating structured strategies that scale with long-context models.
(2) \textbf{A-Mem}~\citep{xu2025amem}: dynamically organizes agent memories using Zettelkasten-style indexing and linking, continuously refining contextual representations as new experience is integrated.
(3) \textbf{ReasoningBank}~\citep{ouyang2026reasoningbank}: distills generalizable reasoning strategies from both successful and failed trajectories into structured memory items.
(4) \textbf{AutoSkill}~\citep{yang2026autoskill}: extracts reusable skills from interaction experience through a lifecycle of extraction, structured representation, iterative refinement, and versioned maintenance.
(5) \textbf{Harness}: inspired by~\citep{lin2026harness, lin2026ahe}, we implement a harness evolution method that jointly maintains and updates system prompts, skills, and experience memory through reflection and revision.

\paragraph{Implementation Details.}
Our evaluation infrastructure is built on Exgentic~\citep{bandel2026general}, a framework that standardizes communication between heterogeneous agent interfaces and benchmarks.
All self-evolving methods are adapted to operate within this framework under a test-time setting.
For text embedding, we adopt the all-MiniLM-L6-v2 model across all experiments.
We sample $N=50$ tasks from each benchmark, where AppWorld uses the test-challenge split, BFCL uses the multi-turn base split, and Tau2 uses the telecom domain.
Task-level performance is evaluated by the native scoring pipeline of each benchmark.
All judge models and user simulator models required by the benchmarks are unified to GPT-5.4.
To account for ordering effects, we run three random seeds that shuffle task order while keeping the task set fixed.
In the \texttt{Sequential} setting, benchmarks are presented in the order AppWorld $\rightarrow$ BFCL $\rightarrow$ BrowseComp+ $\rightarrow$ HLE $\rightarrow$ SWE $\rightarrow$ Tau2.
Across all three streaming scenarios, the within-benchmark task exposure order is held constant to ensure comparability. 

\begin{table*}
  \centering
  \caption{Main results (\%) of self-evolving methods across three streaming scenarios, averaged over three random seeds. \colorbox{avglowbg}{Red background} indicates scores below the model's vanilla baseline, and \textcolor{msftblue}{\textbf{blue}} marks the best scenario within each self-evolving method.}
  \label{tab:seedavg}
  \resizebox{\textwidth}{!}{%
  \begin{tabular}{l@{\hspace{6pt}}l@{\hspace{6pt}}ccccccc}
    \toprule
    Methods & Mode & AppWorld & BFCL & BrowseComp+ & HLE & SWE & Tau2 & Avg \\
    \midrule
    \textit{GPT-5.4} & Vanilla & 44.6 & 66.0 & 50.0 & 2.0 & 62.0 & 50.0 & 45.8 \\
    \midrule
    \multirow{3}{*}{ACE~\citep{zhang2026ace}}
      & \texttt{Isolated} & 39.1\std{5.9} & 68.0\std{12.2} & 46.7\std{2.3} & 5.3\std{1.2} & 60.0\std{3.5} & 63.3\std{6.1} & \avgbest{47.1} \\
      & \texttt{Sequential}    & 38.9\std{5.9} & 64.0\std{5.3} & 48.7\std{6.1} & 7.3\std{1.2} & 62.7\std{5.0} & 45.3\std{22.3} & \avglow{44.5} \\
      & \texttt{Interleaved}   & 32.9\std{1.4} & 58.7\std{3.1} & 44.7\std{2.3} & 6.0\std{2.0} & 63.3\std{5.0} & 48.7\std{3.1} & \avglow{42.4} \\
    \midrule
    \multirow{3}{*}{A-Mem~\citep{xu2025amem}}
      & \texttt{Isolated} & 41.2\std{9.5} & 68.7\std{2.3} & 52.0\std{7.2} & 8.7\std{3.1} & 65.3\std{1.2} & 50.0\std{36.2} & 47.7 \\
      & \texttt{Sequential}    & 39.9\std{4.5} & 66.7\std{6.4} & 50.7\std{3.1} & 10.0\std{2.0} & 60.0\std{5.3} & 62.7\std{18.9} & 48.3 \\
      & \texttt{Interleaved}   & 37.9\std{4.8} & 65.3\std{3.1} & 52.7\std{4.2} & 8.7\std{2.3} & 62.0\std{3.5} & 76.0\std{21.6} & \avgbest{50.4} \\
    \midrule
    \multirow{3}{*}{ReasoningBank~\citep{ouyang2026reasoningbank}}
      & \texttt{Isolated} & 40.9\std{1.8} & 62.7\std{7.0} & 45.3\std{2.3} & 10.0\std{2.0} & 59.3\std{5.0} & 46.7\std{12.2} & \avglow{44.2} \\
      & \texttt{Sequential}    & 42.9\std{0.6} & 62.7\std{2.3} & 47.3\std{3.1} & 14.7\std{4.6} & 62.7\std{4.6} & 36.0\std{14.4} & \avglow{44.4} \\
      & \texttt{Interleaved}   & 38.6\std{3.5} & 63.3\std{5.0} & 48.7\std{4.2} & 12.0\std{2.0} & 56.7\std{6.4} & 53.3\std{4.2} & \cellcolor{avglowbg}\avgbest{45.4} \\
    \midrule
    \multirow{3}{*}{AutoSkill~\citep{yang2026autoskill}}
      & \texttt{Isolated} & 40.1\std{2.8} & 70.7\std{1.2} & 48.0\std{3.5} & 4.7\std{1.2} & 57.3\std{7.0} & 36.0\std{14.0} & \avglow{42.8} \\
      & \texttt{Sequential}    & 38.2\std{1.6} & 72.0\std{3.5} & 43.3\std{5.0} & 6.7\std{1.2} & 58.0\std{4.0} & 44.0\std{13.1} & \avglow{43.7} \\
      & \texttt{Interleaved}   & 40.0\std{1.7} & 72.0\std{5.3} & 46.0\std{2.0} & 4.7\std{1.2} & 60.7\std{1.2} & 40.0\std{8.7} & \cellcolor{avglowbg}\avgbest{43.9} \\
    \midrule
    \multirow{3}{*}{Harness~\citep{lin2026harness, lin2026ahe}}
      & \texttt{Isolated} & 37.4\std{2.3} & 64.0\std{8.7} & 46.0\std{2.0} & 7.3\std{5.8} & 61.3\std{2.3} & 57.3\std{12.2} & \cellcolor{avglowbg}\avgbest{45.6} \\
      & \texttt{Sequential}    & 35.1\std{1.6} & 66.7\std{2.3} & 45.3\std{1.2} & 6.7\std{1.2} & 57.3\std{6.1} & 54.0\std{5.3} & \avglow{44.2} \\
      & \texttt{Interleaved}   & 37.0\std{4.9} & 65.3\std{5.0} & 50.0\std{5.3} & 6.7\std{1.2} & 57.3\std{2.3} & 46.0\std{5.3} & \avglow{43.7} \\
    \midrule
    \textit{Gemini 3.1 Pro} & Vanilla & 41.8 & 58.0 & 34.0 & 52.0 & 64.0 & 90.0 & 56.6 \\
    \midrule
    \multirow{3}{*}{ACE~\citep{zhang2026ace}}
      & \texttt{Isolated} & 43.2\std{1.9} & 62.7\std{13.3} & 44.0\std{5.3} & 50.0\std{0.0} & 64.0\std{0.0} & 90.7\std{9.5} & 59.1 \\
      & \texttt{Sequential}    & 41.2\std{3.2} & 72.0\std{9.2} & 43.3\std{5.0} & 52.0\std{0.0} & 68.0\std{0.0} & 95.3\std{3.1} & \avgbest{62.0} \\
      & \texttt{Interleaved}   & 40.7\std{2.0} & 52.7\std{13.0} & 44.0\std{6.0} & 50.7\std{1.2} & 64.7\std{1.2} & 96.7\std{3.1} & 58.3 \\
    \midrule
    \multirow{3}{*}{A-Mem~\citep{xu2025amem}}
      & \texttt{Isolated} & 41.2\std{6.7} & 54.0\std{25.0} & 50.0\std{4.0} & 52.0\std{2.0} & 62.0\std{5.3} & 92.7\std{4.6} & \avgbest{58.7} \\
      & \texttt{Sequential}    & 40.4\std{3.5} & 54.7\std{2.3} & 46.0\std{2.0} & 48.7\std{4.2} & 60.7\std{1.2} & 91.3\std{2.3} & 57.0 \\
      & \texttt{Interleaved}   & 40.4\std{2.6} & 56.7\std{16.7} & 47.3\std{7.6} & 50.7\std{1.2} & 60.7\std{2.3} & 90.7\std{6.1} & 57.8 \\
    \midrule
    \multirow{3}{*}{ReasoningBank~\citep{ouyang2026reasoningbank}}
      & \texttt{Isolated} & 42.8\std{3.1} & 54.7\std{4.6} & 50.0\std{5.3} & 48.7\std{4.2} & 62.0\std{4.0} & 95.3\std{2.3} & 58.9 \\
      & \texttt{Sequential}    & 42.0\std{0.8} & 56.0\std{2.0} & 44.0\std{2.0} & 51.3\std{1.2} & 66.0\std{5.3} & 94.7\std{3.1} & 59.0 \\
      & \texttt{Interleaved}   & 44.7\std{1.9} & 48.7\std{3.1} & 52.7\std{4.2} & 52.7\std{2.3} & 67.3\std{3.1} & 94.7\std{2.3} & \avgbest{60.1} \\
    \midrule
    \multirow{3}{*}{AutoSkill~\citep{yang2026autoskill}}
      & \texttt{Isolated} & 43.1\std{2.3} & 62.7\std{3.1} & 46.7\std{4.2} & 51.3\std{3.1} & 64.7\std{5.0} & 83.3\std{5.0} & 58.6 \\
      & \texttt{Sequential}    & 45.5\std{1.5} & 64.7\std{5.8} & 41.3\std{4.2} & 50.0\std{3.5} & 62.7\std{3.1} & 86.7\std{5.0} & 58.5 \\
      & \texttt{Interleaved}   & 44.7\std{1.2} & 65.3\std{1.2} & 41.3\std{6.1} & 50.0\std{2.0} & 66.0\std{2.0} & 87.3\std{2.3} & \avgbest{59.1} \\
    \midrule
    \multirow{3}{*}{Harness~\citep{lin2026harness, lin2026ahe}}
      & \texttt{Isolated} & 39.9\std{5.0} & 74.7\std{3.1} & 42.7\std{4.6} & 52.0\std{3.5} & 68.0\std{4.0} & 90.0\std{5.3} & \avgbest{61.2} \\
      & \texttt{Sequential}    & 41.9\std{1.0} & 54.7\std{23.1} & 35.3\std{2.3} & 50.7\std{2.3} & 64.7\std{4.2} & 91.3\std{4.2} & \avglow{56.4} \\
      & \texttt{Interleaved}   & 40.9\std{2.0} & 74.7\std{4.2} & 38.0\std{16.4} & 51.3\std{2.3} & 63.3\std{3.1} & 90.7\std{6.1} & 59.8 \\
    \midrule
    \textit{Claude Opus 4.7} & Vanilla & 41.2 & 86.0 & 70.0 & 38.0 & 68.0 & 80.0 & 63.9 \\
    \midrule
    \multirow{3}{*}{ACE~\citep{zhang2026ace}}
      & \texttt{Isolated} & 47.5\std{4.2} & 85.3\std{1.2} & 72.0\std{2.0} & 33.3\std{4.2} & 70.7\std{2.3} & 93.3\std{3.1} & \avgbest{67.0} \\
      & \texttt{Sequential}    & 47.5\std{0.9} & 82.7\std{3.1} & 68.7\std{1.2} & 38.7\std{2.3} & 70.7\std{4.2} & 91.3\std{5.0} & 66.6 \\
      & \texttt{Interleaved}   & 46.7\std{0.8} & 80.7\std{4.2} & 68.0\std{0.0} & 34.0\std{7.2} & 70.0\std{3.5} & 72.0\std{8.7} & \avglow{61.9} \\
    \midrule
    \multirow{3}{*}{A-Mem~\citep{xu2025amem}}
      & \texttt{Isolated} & 48.7\std{2.5} & 85.3\std{1.2} & 71.3\std{1.2} & 27.3\std{1.2} & 71.3\std{6.4} & 90.7\std{7.6} & 65.8 \\
      & \texttt{Sequential}    & 48.1\std{0.4} & 82.7\std{1.2} & 68.0\std{4.0} & 32.7\std{3.1} & 69.3\std{4.2} & 96.0\std{3.5} & \avgbest{66.1} \\
      & \texttt{Interleaved}   & 49.3\std{0.9} & 81.3\std{1.2} & 67.3\std{2.3} & 32.0\std{2.0} & 72.7\std{6.1} & 86.0\std{6.9} & 64.8 \\
    \midrule
    \multirow{3}{*}{ReasoningBank~\citep{ouyang2026reasoningbank}}
      & \texttt{Isolated} & 44.0\std{1.1} & 84.0\std{0.0} & 70.0\std{2.0} & 36.0\std{2.0} & 69.3\std{4.2} & 90.0\std{0.0} & 65.6 \\
      & \texttt{Sequential}    & 41.8\std{1.4} & 85.3\std{1.2} & 71.3\std{3.1} & 35.3\std{3.1} & 66.7\std{3.1} & 85.3\std{1.2} & 64.3 \\
      & \texttt{Interleaved}   & 45.1\std{0.9} & 84.7\std{1.2} & 70.0\std{3.5} & 39.3\std{3.1} & 68.7\std{1.2} & 88.7\std{3.1} & \avgbest{66.1} \\
    \midrule
    \multirow{3}{*}{AutoSkill~\citep{yang2026autoskill}}
      & \texttt{Isolated} & 41.9\std{1.3} & 86.0\std{0.0} & 73.3\std{1.2} & 36.7\std{1.2} & 68.0\std{5.3} & 82.0\std{3.5} & 64.7 \\
      & \texttt{Sequential}    & 41.8\std{0.5} & 86.0\std{0.0} & 72.7\std{2.3} & 37.3\std{2.3} & 63.3\std{1.2} & 86.7\std{7.6} & 64.6 \\
      & \texttt{Interleaved}   & 40.6\std{2.4} & 86.7\std{1.2} & 73.3\std{1.2} & 38.0\std{3.5} & 67.3\std{3.1} & 86.7\std{2.3} & \avgbest{65.4} \\
    \midrule
    \multirow{3}{*}{Harness~\citep{lin2026harness, lin2026ahe}}
      & \texttt{Isolated} & 44.8\std{1.0} & 84.0\std{2.0} & 68.7\std{1.2} & 34.0\std{5.3} & 69.3\std{4.2} & 90.7\std{3.1} & 65.3 \\
      & \texttt{Sequential}    & 42.6\std{0.8} & 82.7\std{2.3} & 71.3\std{4.2} & 36.0\std{2.0} & 72.0\std{0.0} & 74.0\std{28.0} & \avglow{63.1} \\
      & \texttt{Interleaved}   & 46.1\std{2.1} & 85.3\std{1.2} & 69.3\std{1.2} & 37.3\std{1.2} & 69.3\std{2.3} & 87.3\std{9.2} & \avgbest{65.8} \\
    \bottomrule
  \end{tabular}%
  }
\end{table*}

\section{Results}
\cref{tab:seedavg} reports the main results across models, self-evolving methods and streaming scenarios, averaged over three random seeds. Per-seed results are provided in \cref{tab:seed42,tab:seed44,tab:seed46}.

\subsection{Does Self-Evolution Help at All, and Under Which Scenario?}

Marginalizing over both models and methods in \cref{tab:scenario_analysis}, we find that self-evolution is not uniformly beneficial and underperforms the vanilla baseline in a substantial share of configurations.
Given this variability, we read \cref{tab:seedavg} along the streaming-scenario axis alone, asking under which scenario self-evolution most reliably attains a positive evolution gain.

\begin{table}[tbp]
\vspace{-1.5\baselineskip}
  \centering
  \small
  \caption{Positive rate and average evolution gain (\%) over vanilla baseline for each streaming scenario, aggregated over all model-method configurations. Top-1 rate is the percentage of configurations for which the scenario attains the highest accuracy. \textcolor{msftblue}{\textbf{Blue}} marks the best scenario per metric.}
  \label{tab:scenario_analysis}
  \begin{tabular}{lcccc}
    \toprule
    Scenario & Positive ($>$ Vanilla) & Negative ($<$ Vanilla) & Gain & Top-1 rate \\
    \midrule
    \texttt{Isolated}    &\avgbest{34/45}&\avgbest{11/45}& \avgbest{+1.37}\std{0.80} & \avgbest{38\%} \\
    \texttt{Sequential}  & 28/45 & 16/45 & +0.75\std{0.48}& 29\% \\
    \texttt{Interleaved} & 28/45 & 17/45 & +0.90\std{0.34} & 33\% \\
    \bottomrule
  \end{tabular}%
\end{table}

\paragraph{\texttt{Isolated} is the most reliable streaming scenario for self-evolution.}
As shown in \cref{tab:scenario_analysis}, \texttt{Isolated} achieves the highest positive rate of 75.6\% and the largest average evolution gain of +1.37\%, substantially outperforming both \texttt{Sequential} at 62.2\% positive rate and +0.75\% gain, and \texttt{Interleaved} at 62.2\% and +0.90\%.
This advantage is further validated by its highest Top-1 rate of 38\%.
When all tasks come from a single benchmark, the experience stream remains distributionally coherent. 
Each solved task contributes directly relevant knowledge for subsequent tasks, and no filtering or retrieval gating is needed to avoid cross-domain interference. 
This makes \texttt{Isolated} the lowest-risk streaming setting regardless of method architecture.

\paragraph{The two cross-domain scenarios do not follow the expected difficulty ordering.}
A natural hypothesis is that \texttt{Interleaved}, which exposes the agent to tasks from all domains in mixed order, should pose the greatest challenge due to maximal cross-domain interference. 
However, the empirical results contradict this expectation. Although both scenarios achieve comparable positive rates of 62.3\% in \cref{tab:scenario_analysis}, \texttt{Interleaved} attains a higher average evolution gain of +0.90\% against +0.75\% for \texttt{Sequential}, together with a higher Top-1 rate of 33\% versus 29\%.
The advantage of \texttt{Interleaved} becomes more apparent in direct pairwise comparison. 
Across all 15 configurations in \cref{tab:seedavg}, \texttt{Interleaved} achieves higher average accuracy than \texttt{Sequential} in 10 cases while \texttt{Sequential} leads in only 5. 
This pattern holds across models, with \texttt{Interleaved} prevailing in 4 out of 5 configurations on Gemini 3.1 Pro and 3 out of 5 on Claude Opus 4.7.
On GPT-5.4, where both scenarios yield negative evolution gains, \texttt{Interleaved} still leads in 3 out of 5 configurations. 
Beyond its higher evolution gain, \texttt{Interleaved} also exhibits a smaller standard deviation than \texttt{Sequential}, indicating that its advantage is not only larger but also more stable.
Together, these results suggest that the diversity of interleaved streams compensates for cross-domain noise more effectively than the ordered domain transitions in \texttt{Sequential}.

\begin{findingbox}
Self-evolution is not uniformly beneficial, and its reliability depends on the streaming scenario.
\texttt{Isolated} is the most reliable streaming scenario due to distributional coherence within a single domain. Among the two cross-domain scenarios, \texttt{Interleaved} generally outperforms \texttt{Sequential} despite introducing greater task-level interference.
\end{findingbox}

\subsection{When Does Self-Evolution Help? The Role of Model Capability}
\label{sec:model_setting}
\begin{figure*}[!htbp]
\vspace{-1.5\baselineskip}
    \centering
    \includegraphics[width=1.0\textwidth]{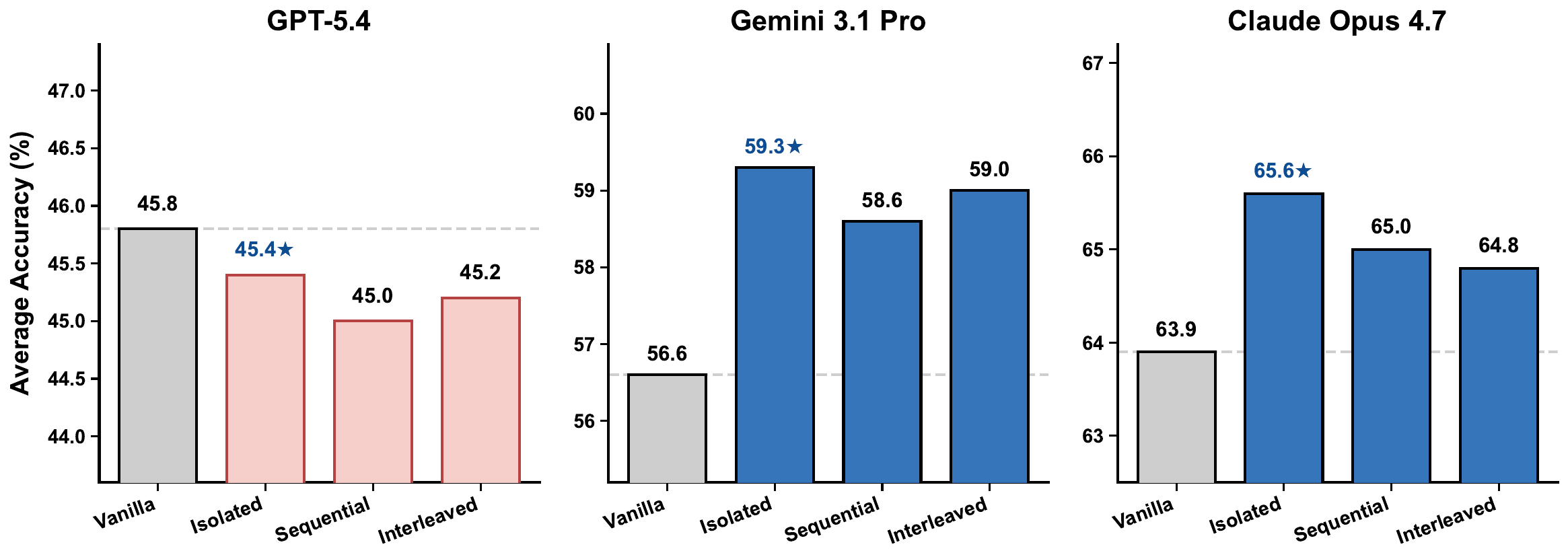}  
    \caption{Average accuracy for each model under the three streaming scenarios.}
    \label{fig:model_mode_margin}
\end{figure*}

Different models may respond differently to self-evolution. \cref{fig:model_mode_margin} provides the marginal view averaged across all five methods, and \cref{tab:model_setting_positiverate} offers a finer-grained breakdown under the three streaming scenarios.

\begin{table}[!htbp]
  \centering
  \small
  \caption{Positive rate and average evolution gain (\%) over vanilla baseline for each model under each streaming scenario. Positive denotes the number of methods, out of 5, exceeding vanilla baseline.}
  \label{tab:model_setting_positiverate}
  \begin{tabular}{lcccccc}
    \toprule
    & \multicolumn{2}{c}{\texttt{Isolated}} & \multicolumn{2}{c}{\texttt{Sequential}} & \multicolumn{2}{c}{\texttt{Interleaved}} \\
    \cmidrule(lr){2-3} \cmidrule(lr){4-5} \cmidrule(lr){6-7}
    Model & Positive & Gain & Positive& Gain & Positive& Gain \\
    \midrule
    GPT-5.4         & 2/5 & $-$0.35 & 1/5 & $-$0.78 & 1/5 & $-$0.62 \\
    Gemini 3.1 Pro  & 5/5 & +2.71 & 4/5 & +1.98 & 5/5 & +2.41 \\
    Claude Opus 4.7 & 5/5 & +1.75 & 4/5 & +1.05 & 4/5 & +0.90 \\
    \bottomrule
  \end{tabular}
  \vspace{-0.5\baselineskip}
\end{table}

\paragraph{The benefit of self-evolution is gated by model capability.}
As shown in \cref{tab:model_setting_positiverate}, GPT-5.4 exhibits negative average evolution gains under all streaming scenarios, ranging from $-$0.35\% to $-$0.78\%, with only 4 out of 15 configurations exceeding its vanilla baseline.
In contrast, Gemini 3.1 Pro exceeds the vanilla baseline in 14 of 15 configurations with gains between +1.98\% and +2.71\%, and Claude Opus 4.7 in 13 of 15 with gains between +0.90\% and +1.75\%.
This disparity persists across \texttt{Isolated}, \texttt{Sequential}, and \texttt{Interleaved} scenarios, as confirmed by per-seed results in \cref{tab:seed42,tab:seed44,tab:seed46}.
These results suggest that self-evolution relies on a bootstrap loop in which successful task completions generate high-quality experience that benefits subsequent tasks.
When the base model's solve rate is low, the experience stream is dominated by failed or partially correct trajectories, from which the agent cannot reliably extract transferable knowledge.
Above a sufficient capability threshold, the model solves enough early tasks to seed the experience buffer with useful patterns, potentially initiating a virtuous cycle of accumulation and transfer that persists across all streaming scenarios.

\paragraph{The evolution gain is non-monotonic in base model strength across streaming scenarios.}
Between the two models that benefit from self-evolution, Gemini 3.1 Pro obtains a larger average evolution gain of +2.37\% than Claude Opus 4.7 at +1.23\%, despite Claude being the stronger model on the vanilla baseline.
As shown in \cref{tab:model_setting_positiverate}, this pattern holds across all streaming scenarios and is corroborated in \cref{tab:seed42,tab:seed44,tab:seed46}.
Moreover, the magnitude of this difference varies with streaming complexity. The gap in evolution gain between Gemini 3.1 Pro and Claude Opus 4.7 is 0.96\% under \texttt{Isolated} and 0.93\% under \texttt{Sequential}, but widens to 1.51\% under \texttt{Interleaved}.
As the streaming scenario introduces more cross-domain mixing, the mid-capability model benefits increasingly more than the stronger model. 
Under \texttt{Interleaved}, the mixed task stream exposes the agent to diverse cross-domain experience, providing substantial learning signals for a model that still has broad room to improve across multiple benchmarks. 
A model that already achieves high baseline accuracy on most benchmarks extracts less additional value from this diversity.

\begin{findingbox}
Self-evolution benefits are gated by model capability: a weak model fails to convert accumulated experience into gains under any streaming scenario. Moreover, the evolution gain is non-monotonic in base model strength, and this effect intensifies with streaming complexity.
\end{findingbox}

\subsection{How Does Model Capability Shape the Choice of Self-Evolving Method?}

Having established that model capability gates when self-evolution helps, we now turn to the interaction between model and self-evolving method, asking how model capability shapes the choice of self-evolving method.
\cref{tab:margin_model_method} reports the accuracy of each method under each model, averaged across the three streaming scenarios.

\paragraph{Method sensitivity decreases with model capability.}
As shown in \cref{tab:margin_model_method}, the spread between the best and worst method contracts monotonically with model strength. On GPT-5.4, the spread is 5.3\% with only 1 of 5 methods exceeding the vanilla baseline, and the worst method losing 2.3\%. On Gemini 3.1 Pro, the spread narrows to 2.0\% with all 5 methods above baseline. On Claude Opus 4.7, it further contracts to 0.9\%, with all methods clustered in a narrow band above the vanilla baseline. Results in \cref{tab:seed42,tab:seed44,tab:seed46} confirm this pattern, with GPT-5.4 exhibiting spreads of 7.0\%, 5.6\%, and 7.7\% against 2.3\%, 3.3\%, and 2.2\% for Claude Opus 4.7.
The trend follows directly from the capability-gating effect identified in \cref{sec:model_setting}. On a weaker model, method architecture determines whether the agent can extract usable signal from a largely unsuccessful experience stream, making method choice decisive. On a stronger model, all methods reliably generate positive transfer, leaving less room for method choice to affect the outcome.

\begin{table}[!htbp]
  \centering
  \small
  \caption{Average accuracy (\%) of each self-evolving method under each model, averaged across three streaming scenarios. \colorbox{avglowbg}{Red background} marks performance below the model's vanilla baseline, and \textcolor{msftblue}{\textbf{blue}} marks the best method within each model.}
  \label{tab:margin_model_method}
  \begin{tabular}{lcccccc}
    \toprule
    Model & Vanilla & ACE & A-Mem & ReasoningBank & AutoSkill & Harness \\
    \midrule
    GPT-5.4         & 45.8 & \avglow{44.6} & \avgbest{48.8} & \avglow{44.7} & \avglow{43.5} & \avglow{44.5} \\
    Gemini 3.1 Pro  & 56.6 & \avgbest{59.8} & 57.8 & 59.4 & 58.8 & 59.2 \\
    Claude Opus 4.7 & 63.9 & 65.2 & \avgbest{65.6} & 65.3 & 64.9 & 64.7 \\
    \bottomrule
  \end{tabular}
\end{table}

\paragraph{Method choice governs whether self-evolution equalizes or amplifies inter-model gaps.}
Because method sensitivity concentrates on the weakest model, the choice of self-evolving method determines inter-model performance gaps. Averaging across all five methods, the gap between GPT-5.4 and Claude Opus 4.7 widens from 18.1\% at vanilla to 19.9\% after evolution, as evolution harms the weakest model on average while consistently benefiting the strongest. However, pairing each model with its optimal method narrows the gap to 16.8\%, as A-Mem~\citep{xu2025amem} lifts GPT-5.4 to 48.8\% while still reaching 65.6\% on Claude Opus 4.7.
Critically, the optimal method is not portable across models. A-Mem is the best method on GPT-5.4 but ranks lowest on Gemini 3.1 Pro, while ACE~\citep{zhang2026ace} leads on Gemini yet ranks third on the other two models. Only ReasoningBank~\citep{ouyang2026reasoningbank} remains competitive across all three models. 

\begin{findingbox}
Method sensitivity is inversely related to model capability. The optimal method varies across models rather than transferring between them, making per-model method selection critical for maximizing self-evolution gains across diverse streaming scenarios.
\end{findingbox}

\subsection{How Do Self-Evolving Methods Interact with Streaming Scenarios?}

\begin{table}[!htbp]
  \centering
  \footnotesize
  \caption{Average evolution gain (\%) of self-evolving methods across three streaming scenarios, aggregated over all models. The last two columns indicate the number of configurations (out of 9) in which \texttt{Isolated} or \texttt{Interleaved} attains higher performance.}
  \label{tab:method_scenario}
  \begin{tabular}{l@{\hspace{6pt}}ccccc}
    \toprule
    Method & \texttt{Isolated} & \texttt{Sequential} & \texttt{Interleaved} & \texttt{Isolated} Higher & \texttt{Interleaved} Higher \\
    \midrule
    ACE& +2.28 & +2.26 & $-$1.26 & 7 & 2 \\
    Harness& +1.91 & $-$0.86 & +1.01 & 5 & 3 \\
    A-Mem& +1.93 & +1.71 & +2.22 & 5 & 4 \\
    ReasoningBank& +0.78 & +0.46 & +1.79 & 2 & 7 \\
    AutoSkill& $-$0.07 & +0.19 & +0.72 & 2 & 7 \\
    \bottomrule
  \end{tabular}
\end{table}

\begin{table*}[!htbp]
  \centering
  \caption{Average accuracy (\%) of each self-evolving method across three streaming scenarios and three models. \textcolor{msftblue}{\textbf{Blue}} marks the best scenario within each self-evolving method.}
  \label{tab:method_scenario2}
  \resizebox{\textwidth}{!}{%
  \begin{tabular}{l@{\hspace{6pt}}ccc ccc ccc}
    \toprule
    & \multicolumn{3}{c}{GPT-5.4} & \multicolumn{3}{c}{Gemini 3.1 Pro} & \multicolumn{3}{c}{Claude Opus 4.7} \\
    \cmidrule(lr){2-4} \cmidrule(lr){5-7} \cmidrule(lr){8-10}
    Method & \texttt{Isolated} & \texttt{Sequential} & \texttt{Interleaved} & \texttt{Isolated} & \texttt{Sequential} & \texttt{Interleaved} & \texttt{Isolated} & \texttt{Sequential} & \texttt{Interleaved} \\
    \midrule
    ACE& \avgbest{47.1} & 44.5 & 42.4 & 59.1 & \avgbest{62.0} & 58.3 & \avgbest{67.0} & 66.6 & 61.9 \\
    Harness& \avgbest{45.6} & 44.2 & 43.7 & \avgbest{61.2} & 56.4 & 59.8 & 65.2 & 63.1 & \avgbest{65.8} \\
    A-Mem& 47.6 & 48.3 & \avgbest{50.4} & \avgbest{58.7} & 57.0 & 57.7 & 65.8 & \avgbest{66.1} & 64.8 \\
    ReasoningBank& 44.2 & 44.4 & \avgbest{45.4} & 58.9 & 59.0 & \avgbest{60.1} & 65.5 & 64.3 & \avgbest{66.1} \\
    AutoSkill& 42.8 & 43.7 & \avgbest{43.9} & 58.6 & 58.5 & \avgbest{59.1} & 64.7 & 64.6 & \avgbest{65.4} \\
    \bottomrule
  \end{tabular}%
  }
\end{table*}

Turning from the model to the streaming scenario, we ask how self-evolving methods interact with the streaming structure.
A single design choice splits self-evolving methods along the scenario axis: how tightly each binds experience to the execution context.
Context-integrated methods such as ACE~\citep{zhang2026ace} and Harness~\citep{lin2026harness, lin2026ahe} fold experience directly into the agent prompt, whereas retrieval-based methods such as ReasoningBank~\citep{ouyang2026reasoningbank}, AutoSkill~\citep{yang2026autoskill}, and A-Mem~\citep{xu2025amem} keep it in an external store and inject only the entries retrieved for the current task.
As shown in \cref{tab:method_scenario}, the two families diverge along this axis. 
Context-integrated methods perform best under \texttt{Isolated}, with ACE attaining the highest average evolution gain of +2.28\%, whereas retrieval-based methods each peak under \texttt{Interleaved}, with A-Mem reaching +2.22\%.
\cref{tab:method_scenario2} corroborates this at the per-model level, where ACE and Harness reach their peak accuracy under \texttt{Isolated} for the majority of model configurations while ReasoningBank and AutoSkill consistently peak under \texttt{Interleaved}.

This interaction between method and streaming scenario can be explained by how each method couples stored knowledge with the execution context.
For context-integrated methods such as ACE and Harness, under \texttt{Isolated}, this tight coupling converts experience into precise domain-specific strategies that transfer reliably across tasks within the same distribution.
When the stream spans multiple domains in \texttt{Interleaved}, the gain of ACE drops sharply from +2.28\% to $-$1.26\%, and Harness decreases from +1.91\% to +1.01\%, indicating that tightly coupled experience is susceptible to cross-domain interference.
For retrieval-based methods such as ReasoningBank, AutoSkill, and A-Mem, this gating mechanism suppresses cross-domain interference under \texttt{Interleaved} by activating only task-relevant experience, while the diversity of the interleaved tasks simultaneously drives the consolidation of transferable patterns across domains.
Under \texttt{Sequential}, these methods maintain moderate gains of +0.46\%, +0.19\%, and +1.71\% respectively, as the retrieval gate filters out experience from earlier domains that is no longer relevant to the current one.

\begin{findingbox}
The optimal self-evolving method depends on the streaming scenario. Context-integrated methods benefit most from \texttt{Isolated} where domain-specific experience accumulates without interference, while retrieval-based methods thrive under \texttt{Interleaved}, where selective retrieval shields them from cross-domain noise and diverse tasks consolidate transferable knowledge.
\end{findingbox}

\section{Conclusion}
We introduce AgentStream, a unified framework that organizes agentic benchmarks into a configurable task stream and evaluates self-evolving agents under three instantiated streaming scenarios across multiple models and methods.
Our analysis reveals that self-evolution reliability varies across streaming scenarios, the benefit of self-evolution is gated by model capability and is non-monotonic in model strength, and no single self-evolving method consistently dominates across models and streaming scenarios.
These results provide practical guidance for deploying self-evolving agents and highlight the value of evaluating them under realistic streaming settings.

\section{Limitations}
Because a given model performs unevenly across benchmarks and our evaluation is instantiated within the Exgentic~\citep{bandel2026general} framework, the notion of model capability strength used throughout this work is grounded in the empirical observations under our specific experimental setup rather than a universally valid ranking of the models.
Different agent frameworks, prompting strategies, or benchmark selections may alter the relative ordering, and our conclusions regarding capability gating should therefore be interpreted within this scope.
In addition, we instantiate the framework with three streaming scenarios and six agentic benchmarks, which cover representative but not exhaustive stream compositions and task domains.
Broader coverage of streaming scenarios, benchmarks, models, and self-evolving methods is a direction for future work, and AgentStream is designed to be extensible along all of these axes.

\bibliographystyle{unsrtnat}
\bibliography{references}

\newpage
\appendix

\section{Cost Analysis}
\begin{figure*}[!htbp]
    \centering
    \includegraphics[width=1.0\textwidth]{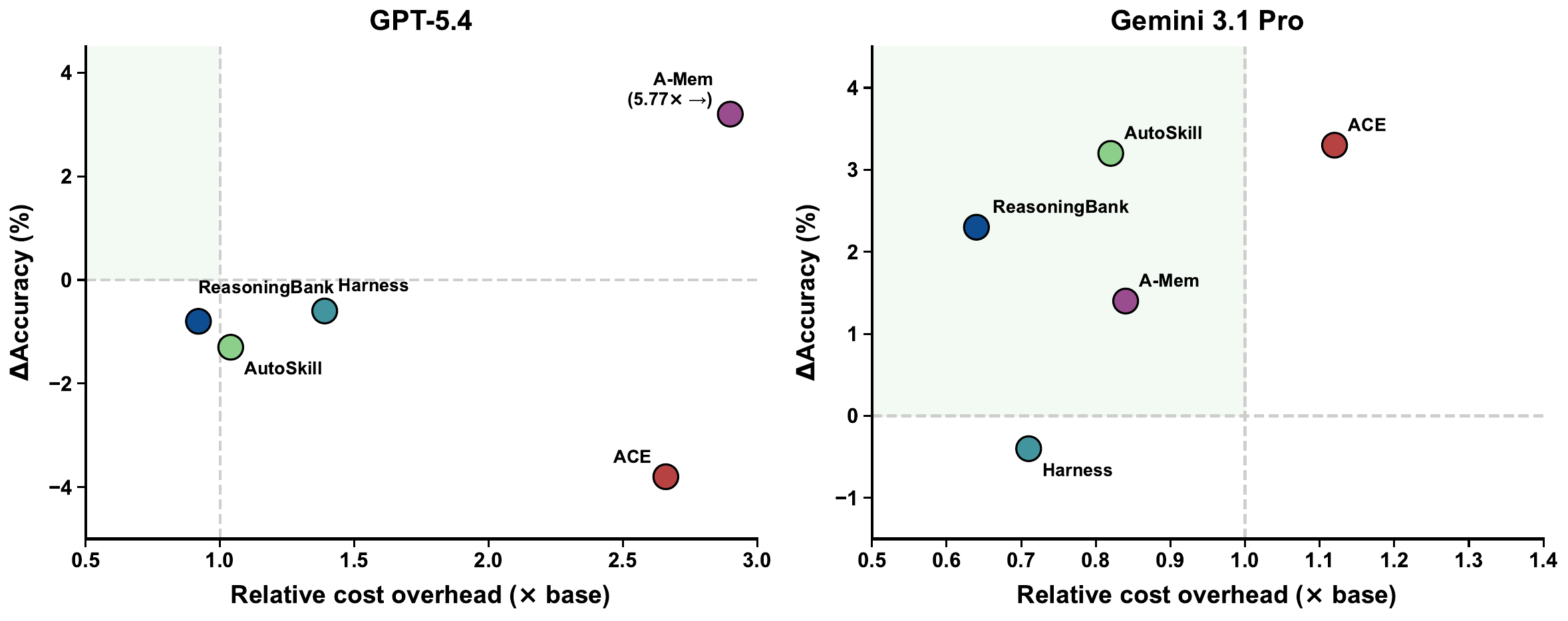}  
    \caption{Cost overhead and evolution gain of each self-evolving method averaged across three streaming scenarios based on a single evaluation.}
    \label{fig:cost_summary}
\end{figure*}

A practical concern for deploying self-evolving agents is whether performance gains justify the additional cost. 
\cref{tab:cost_gpt,tab:cost_gemini} report the per-task cost, relative cost overhead, and average agent steps for GPT-5.4 and Gemini 3.1 Pro respectively, where all costs are reported using LiteLLM's pricing data\footnote{\href{https://github.com/BerriAI/litellm/blob/main/model_prices_and_context_window.json}{Model prices.}}. 
\cref{fig:cost_summary} summarizes these results at the method level, averaged across three streaming scenarios.

\textbf{Self-evolution does not necessarily increase inference cost.} 
On Gemini 3.1 Pro, four methods operate below the vanilla baseline cost, reducing it to 64\% for ReasoningBank~\citep{ouyang2026reasoningbank}, 71\% for Harness~\citep{lin2026harness, lin2026ahe}, 82\% for AutoSkill~\citep{yang2026autoskill}, and 84\% for A-Mem~\citep{xu2025amem}. 
These methods simultaneously reduce the average number of agent steps, suggesting that accumulated experience helps the agent reach solutions more efficiently. 
On GPT-5.4, the pattern reverses. 
Most methods incur higher cost, with A-Mem reaching 577\% of the baseline and ACE~\citep{zhang2026ace} at 266\%, while ReasoningBank remains the only cost-reducing method at 92\% of the baseline.

\textbf{The cost-performance tradeoff is governed by the model.}
On Gemini 3.1 Pro, ACE and AutoSkill both achieve evolution gains above +3\% with moderate or reduced cost overhead. 
ReasoningBank provides +2.3\% evolution gain while reducing cost to 64\% of the baseline, making it the most cost-efficient method on this model.
On GPT-5.4, only A-Mem achieves a positive evolution gain of +3.2\%, but at 577\% of the baseline cost. 
All other methods on GPT-5.4 incur negative evolution gains regardless of their cost overhead. 
This asymmetry echoes the finding in \cref{sec:model_setting} that weaker models struggle to convert accumulated experience into performance improvements, and further shows that this limitation extends to cost efficiency.

\begin{table*}[!htbp]
  \centering
  \footnotesize
  \caption{Per-task cost (\$), relative cost overhead, evolution gain (\%), and average agent steps for GPT-5.4 based on a single evaluation. The vanilla baseline operates at \$0.297 per task with 9.5 steps and 45.8\% accuracy.}
  \label{tab:cost_gpt}
  \begin{tabular}{llccccc}
    \toprule
    Method & Mode & Cost & Relative & $\Delta$Acc. & Steps & $\Delta$Steps (\%) \\
    \midrule
    \multirow{3}{*}{ACE}
      & \texttt{Isolated} & 0.504 & 1.70$\times$ & $-$1.6 & 11.3 & +18.9 \\
      & \texttt{Sequential}    & 0.977 & 3.29$\times$ & $-$7.3 & 10.4 & +9.5 \\
      & \texttt{Interleaved}   & 0.890 & 3.00$\times$ & $-$2.4 & 10.5 & +10.5 \\
    \midrule
    \multirow{3}{*}{Harness}
      & \texttt{Isolated} & 0.385 & 1.30$\times$ & +2.5 & 10.2 & +7.4 \\
      & \texttt{Sequential}    & 0.460 & 1.55$\times$ & $-$2.7 & 10.5 & +10.5 \\
      & \texttt{Interleaved}   & 0.390 & 1.31$\times$ & $-$1.7 & 10.1 & +6.3 \\
    \midrule
    \multirow{3}{*}{A-Mem}
      & \texttt{Isolated} & 1.419 & 4.78$\times$ & $-$2.2 & 8.1 & $-$14.7 \\
      & \texttt{Sequential}  & 1.827 & 6.15$\times$ & +6.2 & 8.6 & $-$9.5 \\
      & \texttt{Interleaved}  & 1.893 & 6.37$\times$ & +5.6 & 8.8 & $-$7.4 \\
    \midrule
    \multirow{3}{*}{ReasoningBank}
      & \texttt{Isolated} & 0.291 & 0.98$\times$ & +1.4 & 8.7 & $-$8.4 \\
      & \texttt{Sequential}    & 0.270 & 0.91$\times$ & $\pm$0.0 & 8.1 & $-$14.7 \\
      & \texttt{Interleaved}   & 0.258 & 0.87$\times$ & $-$3.7 & 8.2 & $-$13.7 \\
    \midrule
    \multirow{3}{*}{AutoSkill}
      & \texttt{Isolated} & 0.298 & 1.00$\times$ & $-$2.3 & 9.1 & $-$4.2 \\
      & \texttt{Sequential}    & 0.324 & 1.09$\times$ & +0.3 & 10.2 & +7.4 \\
      & \texttt{Interleaved}   & 0.309 & 1.04$\times$ & $-$1.8 & 9.4 & $-$1.1 \\
    \bottomrule
  \end{tabular}
\end{table*}

\begin{table*}[!htbp]
  \centering
  \footnotesize
  \caption{Per-task cost (\$), relative cost overhead, evolution gain (\%), and average agent steps for Gemini 3.1 Pro based on a single evaluation. The vanilla baseline operates at \$4.035 per task with 21.1 steps and 56.6\% accuracy.}
  \label{tab:cost_gemini}
  \begin{tabular}{llccccc}
    \toprule
    Method & Mode & Cost & Relative & $\Delta$Acc. & Steps & $\Delta$Steps (\%) \\
    \midrule
    \multirow{3}{*}{ACE}
      & \texttt{Isolated} & 4.239 & 1.05$\times$ & +4.9 & 22.9 & +8.5 \\
      & \texttt{Sequential}    & 3.499 & 0.87$\times$ & +6.6 & 21.0 & $-$0.5 \\
      & \texttt{Interleaved}   & 5.791 & 1.44$\times$ & $-$1.7 & 23.3 & +10.4 \\
    \midrule
    \multirow{3}{*}{Harness}
      & \texttt{Isolated} & 3.791 & 0.94$\times$ & +3.3 & 21.1 & +0.0 \\
      & \texttt{Sequential}    & 2.461 & 0.61$\times$ & $-$4.3 & 18.9 & $-$10.4 \\
      & \texttt{Interleaved}   & 2.310 & 0.57$\times$ & $-$0.2 & 19.3 & $-$8.5 \\
    \midrule
    \multirow{3}{*}{A-Mem}
      & \texttt{Isolated} & 3.508 & 0.87$\times$ & +2.7 & 17.0 & $-$19.4 \\
      & \texttt{Sequential}    & 3.659 & 0.91$\times$ & $-$1.8 & 17.5 & $-$17.1 \\
      & \texttt{Interleaved}   & 2.945 & 0.73$\times$ & +3.4 & 16.2 & $-$23.2 \\
    \midrule
    \multirow{3}{*}{ReasoningBank}
      & \texttt{Isolated} & 2.613 & 0.65$\times$ & +2.5 & 19.7 & $-$6.6 \\
      & \texttt{Sequential}    & 2.680 & 0.66$\times$ & +1.6 & 19.2 & $-$9.0 \\
      & \texttt{Interleaved}   & 2.492 & 0.62$\times$ & +2.9 & 18.8 & $-$10.9 \\
    \midrule
    \multirow{3}{*}{AutoSkill}
      & \texttt{Isolated} & 2.809 & 0.70$\times$ & +2.2 & 20.0 & $-$5.2 \\
      & \texttt{Sequential}    & 3.421 & 0.85$\times$ & +5.0 & 21.0 & $-$0.5 \\
      & \texttt{Interleaved}   & 3.656 & 0.91$\times$ & +2.4 & 22.0 & +4.3 \\
    \bottomrule
  \end{tabular}
\end{table*}

\section{Model Evolution Behavior Analysis}
\label{sec:evolution_behavior}

Beyond aggregate performance, we analyze how the three models differ in their self-evolution behavior under the \texttt{Interleaved} stream, focusing on the evolution state each model accumulates and how that state is updated as tasks are processed.

\subsection{Evolution State Accumulation}

\begin{table}[!htbp]
  \centering
  \small
  \caption{Evolution state accumulated by each model over the \texttt{Interleaved} stream.}
  \label{tab:state_growth}
  \begin{tabular}{llccc}
    \toprule
    Method & Metric & GPT-5.4 & Gemini 3.1 Pro  & Claude Opus 4.7  \\
    \midrule
    ACE & playbook bullets & 552 & 209 & 1251 \\
    ReasoningBank & memory items & 894 & 585 & 898 \\
    A-Mem & evolutions & 261 & 248 & 261 \\
    AutoSkill & skills & 176 & 117 & 160 \\
    Harness & skills & 181 & 137 & 463 \\
    Harness & prompt+memory (chars) & 8,711 & 29,172 & 9,362 \\
    \bottomrule
  \end{tabular}
\end{table}
As shown in \cref{tab:state_growth}, the three models accumulate evolution states of markedly different sizes and structures.
Claude Opus 4.7 generates the largest states across most methods, averaging 1251 ACE~\citep{zhang2026ace} playbook bullets and 463 Harness~\citep{lin2026harness, lin2026ahe} skills, while Gemini 3.1 Pro consistently produces the most compact states with only 209 ACE bullets and 117 AutoSkill~\citep{yang2026autoskill} skills.
GPT-5.4 generates a moderate number of items with relatively concise per-item content, balancing breadth of coverage with manageable state complexity.
The contrast extends beyond volume to architectural preference: under Harness, Claude Opus 4.7 allocates the vast majority of its evolution budget to the skill library while maintaining only 9K characters of system prompt and memory, whereas Gemini 3.1 Pro distributes 29K characters across its system prompt and memory fields, favoring centralized guidance over distributed skill entries.
A similar pattern holds for memory-based methods. Under ReasoningBank~\citep{ouyang2026reasoningbank}, Claude Opus 4.7 and GPT-5.4 each distill approximately 900 memory items while Gemini produces only 585, confirming that Gemini extracts fewer but more consolidated experience entries from each interaction.

\subsection{Update Dynamics}
\cref{tab:update_dynamics} summarizes how the models update their Harness state, which provides the most informative view since it jointly maintains system prompt, memory, and skill library. 
The three models adopt distinct update strategies. GPT-5.4 is largely append-only, adding skills but rarely editing them, with only 5 edits on average over 300 tasks and minimal modification to the system prompt.
Gemini 3.1 Pro spreads its updates more evenly, revising the memory and system prompt 267 times while also refining skills through 66 edits.
Claude Opus 4.7 performs the most intensive refinement, revising on 96\% of tasks while concentrating almost entirely on the skill library rather than the memory or system prompt.
\begin{table}[!htbp]
  \small
  \centering
  \caption{Evolution behavior on Harness state over the \texttt{Interleaved} stream.}
  \label{tab:update_dynamics}
  \begin{tabular}{lccc}
    \toprule
    & GPT-5.4 & Gemini 3.1 Pro  & Claude Opus 4.7 \\
    \midrule
    Revision rate (\%) & 65 & 69 & 96 \\
    Skills added & 181 & 147 & 492 \\
    Skills edited & 5 & 66 & 460 \\
    Memory edited & 20 & 180 & 82 \\
    Prompt edited & 0 & 87 & 21 \\
    \bottomrule
  \end{tabular}
\end{table}

\section{Detailed Results across Random Seeds}
\label{sec:perseed}

The main results in \cref{tab:seedavg} are averaged over three random seeds that share the same task set but differ in the global task arrival order. \cref{tab:seed42,tab:seed44,tab:seed46} provide the complete per-seed results.

\begin{table*}
  \centering
  \footnotesize
  \caption{Main results (\%) of self-evolving methods across three streaming scenarios under seed 42. \colorbox{avglowbg}{Red background} indicates scores below the model's vanilla baseline, and \textcolor{msftblue}{\textbf{blue}} marks the best scenario within each self-evolving method.}
  \label{tab:seed42}
  \begin{tabular}{l@{\hspace{6pt}}l@{\hspace{6pt}}ccccccc}
    \toprule
    Methods & Mode & AppWorld & BFCL & BrowseComp+ & HLE & SWE & Tau2 & Avg \\
    \midrule
    \textit{GPT-5.4} & Vanilla & 44.6 & 66.0 & 50.0 & 2.0 & 62.0 & 50.0 & 45.8 \\
    \midrule
    \multirow{3}{*}{ACE}
      & \texttt{Isolated} & 33.1 & 54.0 & 44.0 & 6.0 & 58.0 & 70.0 & \cellcolor{avglowbg}\avgbest{44.2} \\
      & \texttt{Sequential} & 33.1 & 58.0 & 54.0 & 8.0 & 58.0 & 20.0 & \avglow{38.5} \\
      & \texttt{Interleaved} & 32.3 & 62.0 & 46.0 & 4.0 & 64.0 & 52.0 & \avglow{43.4} \\
    \midrule
    \multirow{3}{*}{A-Mem}
      & \texttt{Isolated} & 41.8 & 70.0 & 60.0 & 12.0 & 66.0 & 12.0 & \avglow{43.6} \\
      & \texttt{Sequential} & 41.8 & 64.0 & 48.0 & 10.0 & 64.0 & 84.0 & \avgbest{52.0} \\
      & \texttt{Interleaved} & 32.4 & 68.0 & 56.0 & 10.0 & 60.0 & 82.0 & 51.4 \\
    \midrule
    \multirow{3}{*}{ReasoningBank}
      & \texttt{Isolated} & 42.9 & 70.0 & 48.0 & 8.0 & 54.0 & 60.0 & \avgbest{47.2} \\
      & \texttt{Sequential} & 42.9 & 60.0 & 48.0 & 12.0 & 60.0 & 52.0 & 45.8 \\
      & \texttt{Interleaved} & 36.6 & 58.0 & 44.0 & 12.0 & 52.0 & 50.0 & \avglow{42.1} \\
    \midrule
    \multirow{3}{*}{AutoSkill}
      & \texttt{Isolated} & 36.8 & 72.0 & 52.0 & 6.0 & 64.0 & 30.0 & \avglow{43.5} \\
      & \texttt{Sequential} & 36.3 & 76.0 & 48.0 & 6.0 & 54.0 & 56.0 & \avgbest{46.1} \\
      & \texttt{Interleaved} & 42.0 & 66.0 & 46.0 & 4.0 & 60.0 & 46.0 & \avglow{44.0} \\
    \midrule
    \multirow{3}{*}{Harness}
      & \texttt{Isolated} & 37.9 & 70.0 & 48.0 & 14.0 & 60.0 & 60.0 & \avgbest{48.3} \\
      & \texttt{Sequential} & 36.4 & 68.0 & 46.0 & 8.0 & 52.0 & 48.0 & \avglow{43.1} \\
      & \texttt{Interleaved} & 40.6 & 66.0 & 46.0 & 8.0 & 60.0 & 44.0 & \avglow{44.1} \\
    \midrule
    \textit{Gemini 3.1 Pro} & Vanilla & 41.8 & 58.0 & 34.0 & 52.0 & 64.0 & 90.0 & 56.6 \\
    \midrule
    \multirow{3}{*}{ACE}
      & \texttt{Isolated} & 41.2 & 74.0 & 46.0 & 50.0 & 64.0 & 94.0 & 61.5 \\
      & \texttt{Sequential} & 40.9 & 74.0 & 48.0 & 52.0 & 68.0 & 96.0 & \avgbest{63.2} \\
      & \texttt{Interleaved} & 41.6 & 40.0 & 38.0 & 52.0 & 64.0 & 94.0 & \avglow{54.9} \\
    \midrule
    \multirow{3}{*}{A-Mem}
      & \texttt{Isolated} & 33.6 & 62.0 & 54.0 & 50.0 & 66.0 & 90.0 & 59.3 \\
      & \texttt{Sequential} & 36.9 & 52.0 & 46.0 & 44.0 & 60.0 & 90.0 & \avglow{54.8} \\
      & \texttt{Interleaved} & 37.9 & 62.0 & 56.0 & 50.0 & 62.0 & 92.0 & \avgbest{60.0} \\
    \midrule
    \multirow{3}{*}{ReasoningBank}
      & \texttt{Isolated} & 44.4 & 60.0 & 44.0 & 50.0 & 62.0 & 94.0 & 59.1 \\
      & \texttt{Sequential} & 41.3 & 58.0 & 42.0 & 52.0 & 64.0 & 92.0 & 58.2 \\
      & \texttt{Interleaved} & 46.7 & 46.0 & 54.0 & 50.0 & 68.0 & 92.0 & \avgbest{59.5} \\
    \midrule
    \multirow{3}{*}{AutoSkill}
      & \texttt{Isolated} & 44.7 & 66.0 & 48.0 & 52.0 & 64.0 & 78.0 & 58.8 \\
      & \texttt{Sequential} & 45.5 & 68.0 & 46.0 & 52.0 & 66.0 & 92.0 & \avgbest{61.6} \\
      & \texttt{Interleaved} & 45.8 & 64.0 & 40.0 & 52.0 & 66.0 & 86.0 & 59.0 \\
    \midrule
    \multirow{3}{*}{Harness}
      & \texttt{Isolated} & 45.5 & 72.0 & 40.0 & 50.0 & 64.0 & 88.0 & \avgbest{59.9} \\
      & \texttt{Sequential} & 41.8 & 28.0 & 34.0 & 52.0 & 68.0 & 90.0 & \avglow{52.3} \\
      & \texttt{Interleaved} & 38.6 & 70.0 & 20.0 & 54.0 & 64.0 & 92.0 & \avglow{56.4} \\
    \midrule
    \textit{Claude Opus 4.7} & Vanilla & 41.2 & 86.0 & 70.0 & 38.0 & 68.0 & 80.0 & 63.9 \\
    \midrule
    \multirow{3}{*}{ACE}
      & \texttt{Isolated} & 48.3 & 86.0 & 70.0 & 32.0 & 68.0 & 90.0 & \avgbest{65.7} \\
      & \texttt{Sequential} & 48.5 & 82.0 & 68.0 & 36.0 & 66.0 & 86.0 & 64.4 \\
      & \texttt{Interleaved} & 47.5 & 82.0 & 68.0 & 26.0 & 68.0 & 78.0 & \avglow{61.6} \\
    \midrule
    \multirow{3}{*}{A-Mem}
      & \texttt{Isolated} & 46.2 & 86.0 & 70.0 & 28.0 & 64.0 & 94.0 & 64.7 \\
      & \texttt{Sequential} & 48.0 & 82.0 & 68.0 & 36.0 & 74.0 & 98.0 & \avgbest{67.7} \\
      & \texttt{Interleaved} & 49.0 & 80.0 & 70.0 & 30.0 & 78.0 & 90.0 & 66.2 \\
    \midrule
    \multirow{3}{*}{ReasoningBank}
      & \texttt{Isolated} & 43.9 & 84.0 & 68.0 & 34.0 & 68.0 & 90.0 & 64.6 \\
      & \texttt{Sequential} & 40.2 & 86.0 & 68.0 & 32.0 & 66.0 & 84.0 & \avglow{62.7} \\
      & \texttt{Interleaved} & 44.1 & 84.0 & 68.0 & 36.0 & 68.0 & 92.0 & \avgbest{65.4} \\
    \midrule
    \multirow{3}{*}{AutoSkill}
      & \texttt{Isolated} & 42.6 & 86.0 & 74.0 & 36.0 & 72.0 & 78.0 & 64.8 \\
      & \texttt{Sequential} & 41.2 & 86.0 & 74.0 & 36.0 & 64.0 & 78.0 & \avglow{63.2} \\
      & \texttt{Interleaved} & 42.9 & 86.0 & 74.0 & 40.0 & 70.0 & 88.0 & \avgbest{66.8} \\
    \midrule
    \multirow{3}{*}{Harness}
      & \texttt{Isolated} & 43.8 & 84.0 & 70.0 & 28.0 & 66.0 & 88.0 & \avglow{63.3} \\
      & \texttt{Sequential} & 42.5 & 84.0 & 68.0 & 34.0 & 72.0 & 94.0 & \avgbest{65.7} \\
      & \texttt{Interleaved} & 45.9 & 84.0 & 68.0 & 38.0 & 68.0 & 82.0 & 64.3 \\
    \bottomrule
  \end{tabular}
\end{table*}

\begin{table*}
  \centering
  \footnotesize
  \caption{Main results (\%) of self-evolving methods across three streaming scenarios under seed 44. \colorbox{avglowbg}{Red background} indicates scores below the model's vanilla baseline, and \textcolor{msftblue}{\textbf{blue}} marks the best scenario within each self-evolving method.}
  \label{tab:seed44}
  \begin{tabular}{l@{\hspace{6pt}}l@{\hspace{6pt}}ccccccc}
    \toprule
    Methods & Mode & AppWorld & BFCL & BrowseComp+ & HLE & SWE & Tau2 & Avg \\
    \midrule
    \textit{GPT-5.4} & Vanilla & 44.6 & 66.0 & 50.0 & 2.0 & 62.0 & 50.0 & 45.8 \\
    \midrule
    \multirow{3}{*}{ACE}
      & \texttt{Isolated} & 44.8 & 76.0 & 48.0 & 6.0 & 64.0 & 62.0 & \avgbest{50.1} \\
      & \texttt{Sequential} & 44.8 & 68.0 & 42.0 & 6.0 & 62.0 & 54.0 & 46.1 \\
      & \texttt{Interleaved} & 34.5 & 58.0 & 46.0 & 8.0 & 58.0 & 48.0 & \avglow{42.1} \\
    \midrule
    \multirow{3}{*}{A-Mem}
      & \texttt{Isolated} & 31.4 & 66.0 & 50.0 & 6.0 & 66.0 & 54.0 & \avglow{45.6} \\
      & \texttt{Sequential} & 34.8 & 62.0 & 54.0 & 12.0 & 62.0 & 48.0 & \avglow{45.5} \\
      & \texttt{Interleaved} & 41.2 & 62.0 & 54.0 & 6.0 & 66.0 & 94.0 & \avgbest{53.9} \\
    \midrule
    \multirow{3}{*}{ReasoningBank}
      & \texttt{Isolated} & 39.7 & 56.0 & 44.0 & 12.0 & 60.0 & 44.0 & \avglow{42.6} \\
      & \texttt{Sequential} & 42.3 & 64.0 & 50.0 & 12.0 & 60.0 & 24.0 & \avglow{42.1} \\
      & \texttt{Interleaved} & 36.6 & 68.0 & 50.0 & 14.0 & 54.0 & 58.0 & \avgbest{46.8} \\
    \midrule
    \multirow{3}{*}{AutoSkill}
      & \texttt{Isolated} & 41.4 & 70.0 & 46.0 & 4.0 & 58.0 & 52.0 & \cellcolor{avglowbg}\avgbest{45.2} \\
      & \texttt{Sequential} & 39.2 & 70.0 & 44.0 & 6.0 & 58.0 & 46.0 & \avglow{43.9} \\
      & \texttt{Interleaved} & 38.7 & 76.0 & 44.0 & 4.0 & 60.0 & 44.0 & \avglow{44.5} \\
    \midrule
    \multirow{3}{*}{Harness}
      & \texttt{Isolated} & 34.9 & 54.0 & 44.0 & 4.0 & 64.0 & 44.0 & \avglow{40.8} \\
      & \texttt{Sequential} & 33.4 & 68.0 & 46.0 & 6.0 & 56.0 & 56.0 & \cellcolor{avglowbg}\avgbest{44.2} \\
      & \texttt{Interleaved} & 39.1 & 60.0 & 56.0 & 6.0 & 56.0 & 42.0 & \avglow{43.2} \\
    \midrule
    \textit{Gemini 3.1 Pro} & Vanilla & 41.8 & 58.0 & 34.0 & 52.0 & 64.0 & 90.0 & 56.6 \\
    \midrule
    \multirow{3}{*}{ACE}
      & \texttt{Isolated} & 43.6 & 66.0 & 38.0 & 50.0 & 64.0 & 80.0 & 56.9 \\
      & \texttt{Sequential} & 44.6 & 80.0 & 44.0 & 52.0 & 68.0 & 92.0 & \avgbest{63.4} \\
      & \texttt{Interleaved} & 38.4 & 52.0 & 50.0 & 50.0 & 66.0 & 100.0 & 59.4 \\
    \midrule
    \multirow{3}{*}{A-Mem}
      & \texttt{Isolated} & 46.3 & 26.0 & 46.0 & 54.0 & 64.0 & 90.0 & \avglow{54.4} \\
      & \texttt{Sequential} & 40.5 & 56.0 & 44.0 & 50.0 & 62.0 & 94.0 & \avgbest{57.8} \\
      & \texttt{Interleaved} & 43.0 & 38.0 & 44.0 & 50.0 & 62.0 & 84.0 & \avglow{53.5} \\
    \midrule
    \multirow{3}{*}{ReasoningBank}
      & \texttt{Isolated} & 39.2 & 52.0 & 52.0 & 52.0 & 66.0 & 94.0 & 59.2 \\
      & \texttt{Sequential} & 41.9 & 56.0 & 46.0 & 50.0 & 62.0 & 94.0 & 58.3 \\
      & \texttt{Interleaved} & 44.5 & 48.0 & 56.0 & 54.0 & 64.0 & 96.0 & \avgbest{60.4} \\
    \midrule
    \multirow{3}{*}{AutoSkill}
      & \texttt{Isolated} & 44.1 & 60.0 & 50.0 & 48.0 & 70.0 & 88.0 & 60.0 \\
      & \texttt{Sequential} & 44.1 & 68.0 & 38.0 & 46.0 & 62.0 & 86.0 & 57.4 \\
      & \texttt{Interleaved} & 43.5 & 66.0 & 48.0 & 48.0 & 68.0 & 90.0 & \avgbest{60.6} \\
    \midrule
    \multirow{3}{*}{Harness}
      & \texttt{Isolated} & 36.0 & 74.0 & 40.0 & 56.0 & 72.0 & 86.0 & \avgbest{60.7} \\
      & \texttt{Sequential} & 43.0 & 68.0 & 38.0 & 52.0 & 66.0 & 96.0 & 60.5 \\
      & \texttt{Interleaved} & 41.9 & 76.0 & 42.0 & 50.0 & 66.0 & 84.0 & 60.0 \\
    \midrule
    \textit{Claude Opus 4.7} & Vanilla & 41.2 & 86.0 & 70.0 & 38.0 & 68.0 & 80.0 & 63.9 \\
    \midrule
    \multirow{3}{*}{ACE}
      & \texttt{Isolated} & 42.9 & 86.0 & 74.0 & 30.0 & 72.0 & 96.0 & 66.8 \\
      & \texttt{Sequential} & 47.1 & 80.0 & 70.0 & 40.0 & 74.0 & 92.0 & \avgbest{67.2} \\
      & \texttt{Interleaved} & 45.9 & 84.0 & 68.0 & 36.0 & 74.0 & 76.0 & 64.0 \\
    \midrule
    \multirow{3}{*}{A-Mem}
      & \texttt{Isolated} & 48.8 & 84.0 & 72.0 & 26.0 & 76.0 & 82.0 & \avgbest{64.8} \\
      & \texttt{Sequential} & 47.8 & 82.0 & 64.0 & 32.0 & 68.0 & 92.0 & 64.3 \\
      & \texttt{Interleaved} & 50.3 & 82.0 & 66.0 & 32.0 & 66.0 & 78.0 & \avglow{62.4} \\
    \midrule
    \multirow{3}{*}{ReasoningBank}
      & \texttt{Isolated} & 43.0 & 84.0 & 70.0 & 36.0 & 66.0 & 90.0 & 64.8 \\
      & \texttt{Sequential} & 43.0 & 86.0 & 74.0 & 38.0 & 64.0 & 86.0 & 65.2 \\
      & \texttt{Interleaved} & 45.6 & 86.0 & 74.0 & 42.0 & 68.0 & 86.0 & \avgbest{66.9} \\
    \midrule
    \multirow{3}{*}{AutoSkill}
      & \texttt{Isolated} & 42.8 & 86.0 & 74.0 & 36.0 & 62.0 & 84.0 & 64.1 \\
      & \texttt{Sequential} & 42.1 & 86.0 & 74.0 & 36.0 & 62.0 & 90.0 & 65.0 \\
      & \texttt{Interleaved} & 40.8 & 88.0 & 74.0 & 40.0 & 68.0 & 84.0 & \avgbest{65.8} \\
    \midrule
    \multirow{3}{*}{Harness}
      & \texttt{Isolated} & 44.9 & 82.0 & 68.0 & 38.0 & 74.0 & 90.0 & \avgbest{66.1} \\
      & \texttt{Sequential} & 41.9 & 80.0 & 70.0 & 36.0 & 72.0 & 42.0 & \avglow{57.0} \\
      & \texttt{Interleaved} & 44.1 & 86.0 & 70.0 & 36.0 & 72.0 & 82.0 & 65.0 \\
    \bottomrule
  \end{tabular}
\end{table*}

\begin{table*}
  \centering
  \footnotesize
  \caption{Main results (\%) of self-evolving methods across three streaming scenarios under seed 46. \colorbox{avglowbg}{Red background} indicates scores below the model's vanilla baseline, and \textcolor{msftblue}{\textbf{blue}} marks the best scenario within each self-evolving method.}
  \label{tab:seed46}
  \begin{tabular}{l@{\hspace{6pt}}l@{\hspace{6pt}}ccccccc}
    \toprule
    Methods & Mode & AppWorld & BFCL & BrowseComp+ & HLE & SWE & Tau2 & Avg \\
    \midrule
    \textit{GPT-5.4} & Vanilla & 44.6 & 66.0 & 50.0 & 2.0 & 62.0 & 50.0 & 45.8 \\
    \midrule
    \multirow{3}{*}{ACE}
      & \texttt{Isolated} & 39.3 & 74.0 & 48.0 & 4.0 & 58.0 & 58.0 & 46.9 \\
      & \texttt{Sequential} & 38.7 & 66.0 & 50.0 & 8.0 & 68.0 & 62.0 & \avgbest{48.8} \\
      & \texttt{Interleaved} & 31.9 & 56.0 & 42.0 & 6.0 & 68.0 & 46.0 & \avglow{41.7} \\
    \midrule
    \multirow{3}{*}{A-Mem}
      & \texttt{Isolated} & 50.3 & 70.0 & 46.0 & 8.0 & 64.0 & 84.0 & \avgbest{53.7} \\
      & \texttt{Sequential} & 43.1 & 74.0 & 50.0 & 8.0 & 54.0 & 56.0 & 47.5 \\
      & \texttt{Interleaved} & 40.0 & 66.0 & 48.0 & 10.0 & 60.0 & 52.0 & 46.0 \\
    \midrule
    \multirow{3}{*}{ReasoningBank}
      & \texttt{Isolated} & 40.0 & 62.0 & 44.0 & 10.0 & 64.0 & 36.0 & \avglow{42.7} \\
      & \texttt{Sequential} & 43.4 & 64.0 & 44.0 & 20.0 & 68.0 & 32.0 & \avglow{45.2} \\
      & \texttt{Interleaved} & 42.6 & 64.0 & 52.0 & 10.0 & 64.0 & 52.0 & \avgbest{47.4} \\
    \midrule
    \multirow{3}{*}{AutoSkill}
      & \texttt{Isolated} & 42.0 & 70.0 & 46.0 & 4.0 & 50.0 & 26.0 & \avglow{39.7} \\
      & \texttt{Sequential} & 39.0 & 70.0 & 38.0 & 8.0 & 62.0 & 30.0 & \avglow{41.2} \\
      & \texttt{Interleaved} & 39.4 & 74.0 & 48.0 & 6.0 & 62.0 & 30.0 & \cellcolor{avglowbg}\avgbest{43.2} \\
    \midrule
    \multirow{3}{*}{Harness}
      & \texttt{Isolated} & 39.5 & 68.0 & 46.0 & 4.0 & 60.0 & 68.0 & \avgbest{47.6} \\
      & \texttt{Sequential} & 35.6 & 64.0 & 44.0 & 6.0 & 64.0 & 58.0 & \avglow{45.3} \\
      & \texttt{Interleaved} & 31.4 & 70.0 & 48.0 & 6.0 & 56.0 & 52.0 & \avglow{43.9} \\
    \midrule
    \textit{Gemini 3.1 Pro} & Vanilla & 41.8 & 58.0 & 34.0 & 52.0 & 64.0 & 90.0 & 56.6 \\
    \midrule
    \multirow{3}{*}{ACE}
      & \texttt{Isolated} & 44.9 & 48.0 & 48.0 & 50.0 & 64.0 & 98.0 & 58.8 \\
      & \texttt{Sequential} & 38.2 & 62.0 & 38.0 & 52.0 & 68.0 & 98.0 & 59.4 \\
      & \texttt{Interleaved} & 42.1 & 66.0 & 44.0 & 50.0 & 64.0 & 96.0 & \avgbest{60.4} \\
    \midrule
    \multirow{3}{*}{A-Mem}
      & \texttt{Isolated} & 43.7 & 74.0 & 50.0 & 52.0 & 56.0 & 98.0 & \avgbest{62.3} \\
      & \texttt{Sequential} & 43.8 & 56.0 & 48.0 & 52.0 & 60.0 & 90.0 & 58.3 \\
      & \texttt{Interleaved} & 40.2 & 70.0 & 42.0 & 52.0 & 58.0 & 96.0 & 59.7 \\
    \midrule
    \multirow{3}{*}{ReasoningBank}
      & \texttt{Isolated} & 44.7 & 52.0 & 54.0 & 44.0 & 58.0 & 98.0 & 58.5 \\
      & \texttt{Sequential} & 42.8 & 54.0 & 44.0 & 52.0 & 72.0 & 98.0 & \avgbest{60.5} \\
      & \texttt{Interleaved} & 43.0 & 52.0 & 48.0 & 54.0 & 70.0 & 96.0 & \avgbest{60.5} \\
    \midrule
    \multirow{3}{*}{AutoSkill}
      & \texttt{Isolated} & 40.4 & 62.0 & 42.0 & 54.0 & 60.0 & 84.0 & 57.1 \\
      & \texttt{Sequential} & 47.0 & 58.0 & 40.0 & 52.0 & 60.0 & 82.0 & \avglow{56.5} \\
      & \texttt{Interleaved} & 44.9 & 66.0 & 36.0 & 50.0 & 64.0 & 86.0 & \avgbest{57.8} \\
    \midrule
    \multirow{3}{*}{Harness}
      & \texttt{Isolated} & 38.3 & 78.0 & 48.0 & 50.0 & 68.0 & 96.0 & \avgbest{63.1} \\
      & \texttt{Sequential} & 41.0 & 68.0 & 34.0 & 48.0 & 60.0 & 88.0 & \avglow{56.5} \\
      & \texttt{Interleaved} & 42.3 & 78.0 & 52.0 & 50.0 & 60.0 & 96.0 & \avgbest{63.1} \\
    \midrule
    \textit{Claude Opus 4.7} & Vanilla & 41.2 & 86.0 & 70.0 & 38.0 & 68.0 & 80.0 & 63.9 \\
    \midrule
    \multirow{3}{*}{ACE}
      & \texttt{Isolated} & 51.2 & 84.0 & 72.0 & 38.0 & 72.0 & 94.0 & \avgbest{68.5} \\
      & \texttt{Sequential} & 46.9 & 86.0 & 68.0 & 40.0 & 72.0 & 96.0 & 68.2 \\
      & \texttt{Interleaved} & 46.6 & 76.0 & 68.0 & 40.0 & 68.0 & 62.0 & \avglow{60.1} \\
    \midrule
    \multirow{3}{*}{A-Mem}
      & \texttt{Isolated} & 51.2 & 86.0 & 72.0 & 28.0 & 74.0 & 96.0 & \avgbest{67.9} \\
      & \texttt{Sequential} & 48.6 & 84.0 & 72.0 & 30.0 & 66.0 & 98.0 & 66.4 \\
      & \texttt{Interleaved} & 48.7 & 82.0 & 66.0 & 34.0 & 74.0 & 90.0 & 65.8 \\
    \midrule
    \multirow{3}{*}{ReasoningBank}
      & \texttt{Isolated} & 45.1 & 84.0 & 72.0 & 38.0 & 74.0 & 90.0 & \avgbest{67.2} \\
      & \texttt{Sequential} & 42.1 & 84.0 & 72.0 & 36.0 & 70.0 & 86.0 & 65.0 \\
      & \texttt{Interleaved} & 45.7 & 84.0 & 68.0 & 40.0 & 70.0 & 88.0 & 66.0 \\
    \midrule
    \multirow{3}{*}{AutoSkill}
      & \texttt{Isolated} & 40.4 & 86.0 & 72.0 & 38.0 & 70.0 & 84.0 & 65.1 \\
      & \texttt{Sequential} & 42.1 & 86.0 & 70.0 & 40.0 & 64.0 & 92.0 & \avgbest{65.7} \\
      & \texttt{Interleaved} & 38.1 & 86.0 & 72.0 & 34.0 & 64.0 & 88.0 & \avglow{63.7} \\
    \midrule
    \multirow{3}{*}{Harness}
      & \texttt{Isolated} & 45.8 & 86.0 & 68.0 & 36.0 & 68.0 & 94.0 & 66.3 \\
      & \texttt{Sequential} & 43.5 & 84.0 & 76.0 & 38.0 & 72.0 & 86.0 & 66.6 \\
      & \texttt{Interleaved} & 48.2 & 86.0 & 70.0 & 38.0 & 68.0 & 98.0 & \avgbest{68.0} \\
    \bottomrule
  \end{tabular}
\end{table*}

\section{Cumulative Accuracy Dynamics}
\label{sec:cumulative_dynamics}
\begin{figure}[!htbp]
    \centering
    \includegraphics[width=0.9\textwidth]{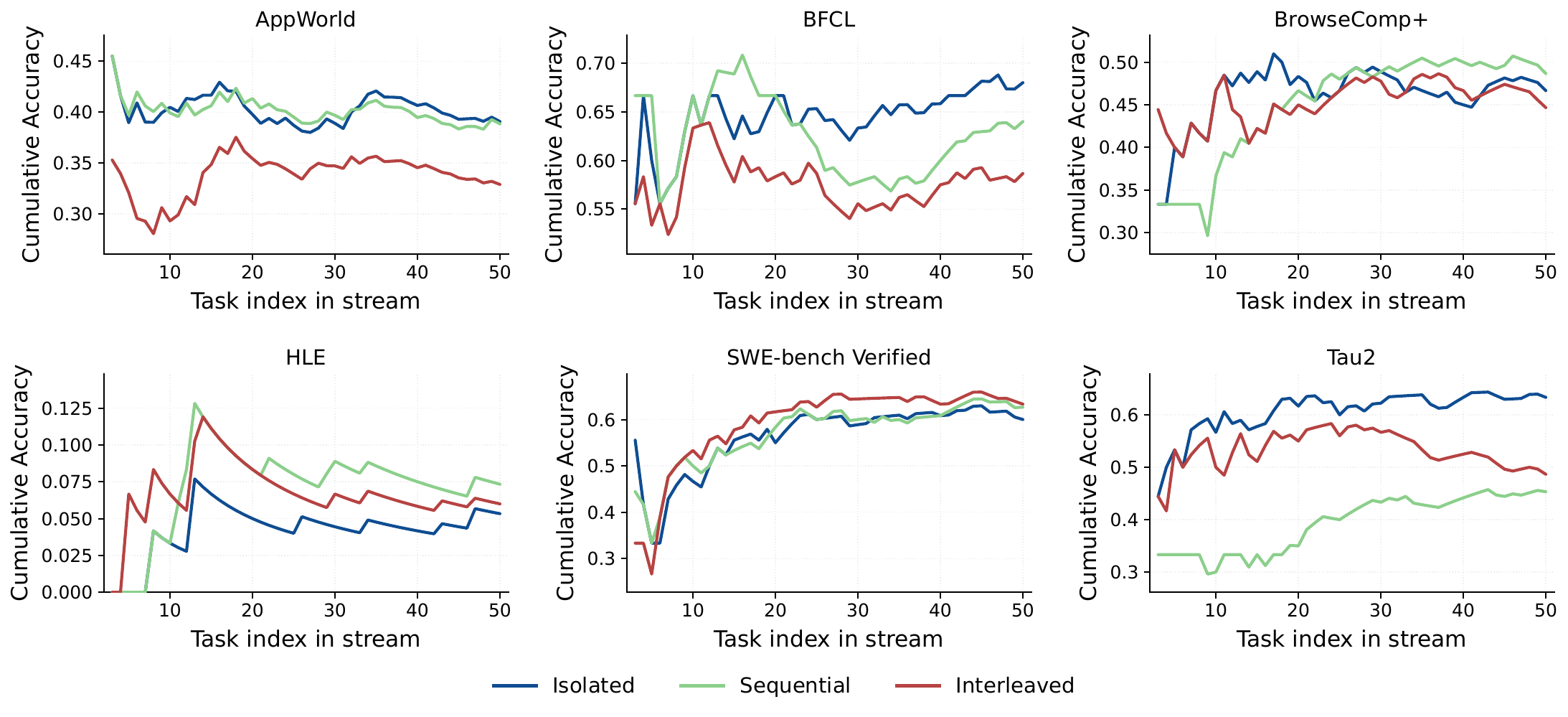}
    \caption{Cumulative accuracy dynamics of ACE on GPT-5.4 over three streaming scenarios.}
    \label{fig:cumulative_ace_gpt}
\end{figure}

\begin{figure}[!htbp]
    \centering
    \includegraphics[width=0.9\textwidth]{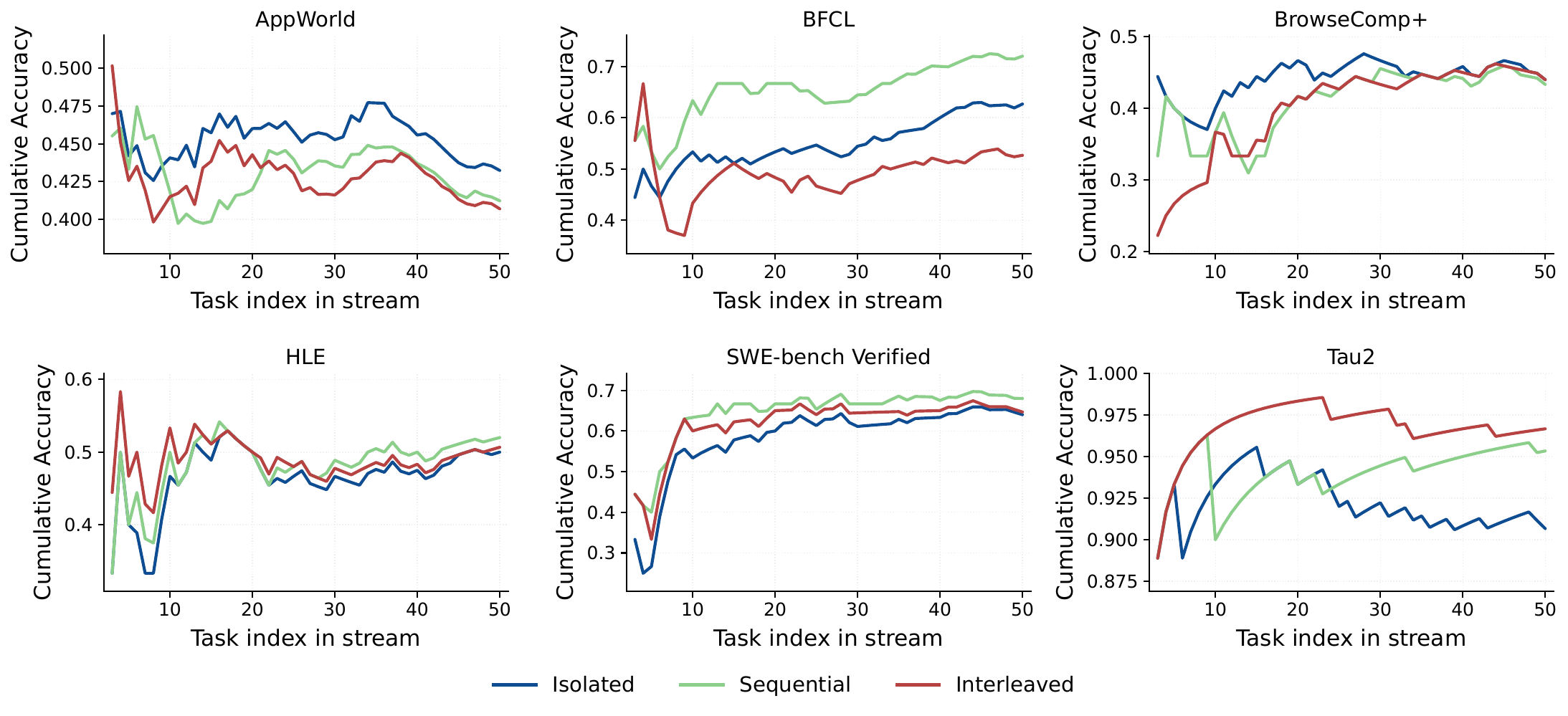}
    \caption{Cumulative accuracy dynamics of ACE on Gemini 3.1 Pro over three streaming scenarios.}
    \label{fig:cumulative_ace_gemini}
\end{figure}

\begin{figure}[!htbp]
    \centering
    \includegraphics[width=0.9\textwidth]{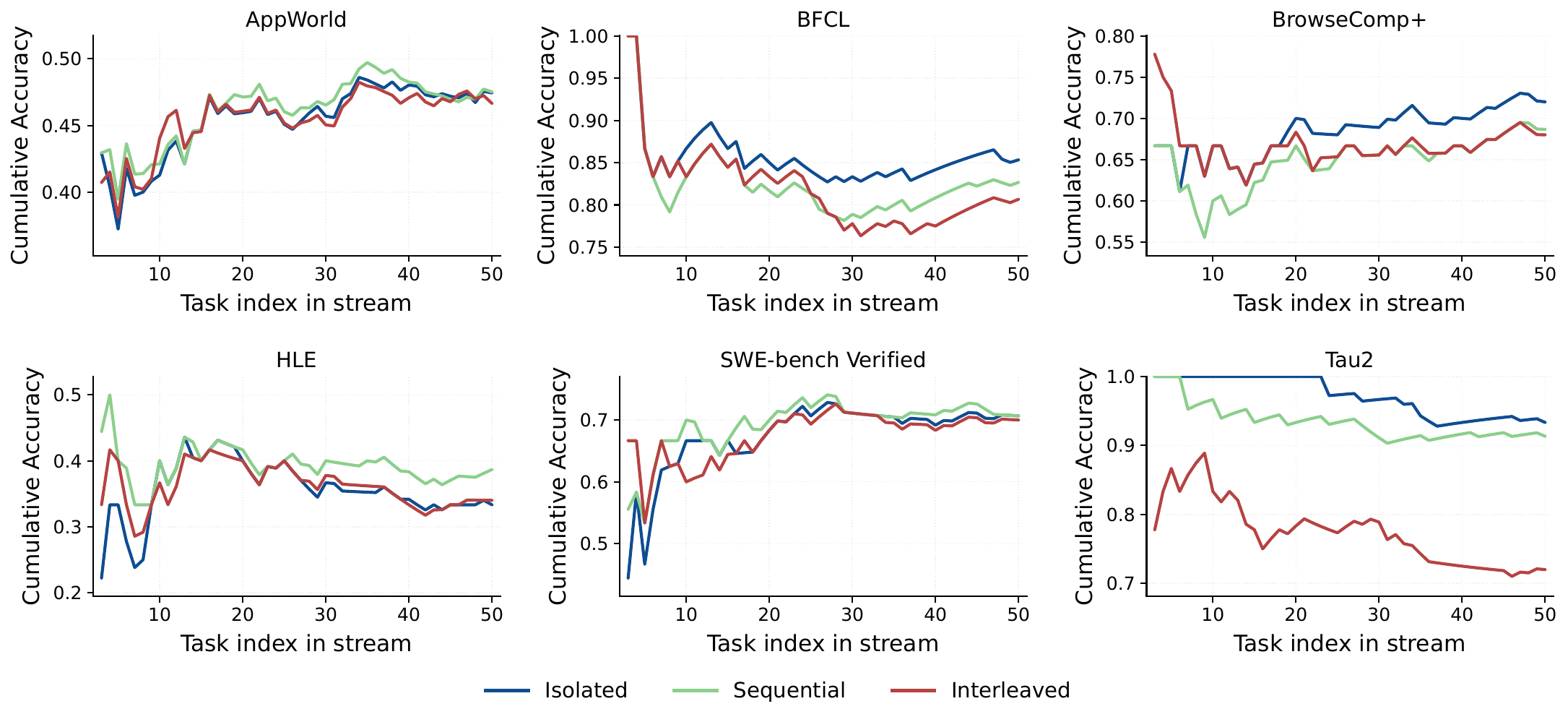}
    \caption{Cumulative accuracy dynamics of ACE on Claude Opus 4.7 over three streaming scenarios.}
    \label{fig:cumulative_ace_claude}
\end{figure}

\begin{figure}[!htbp]
    \centering
    \includegraphics[width=0.9\textwidth]{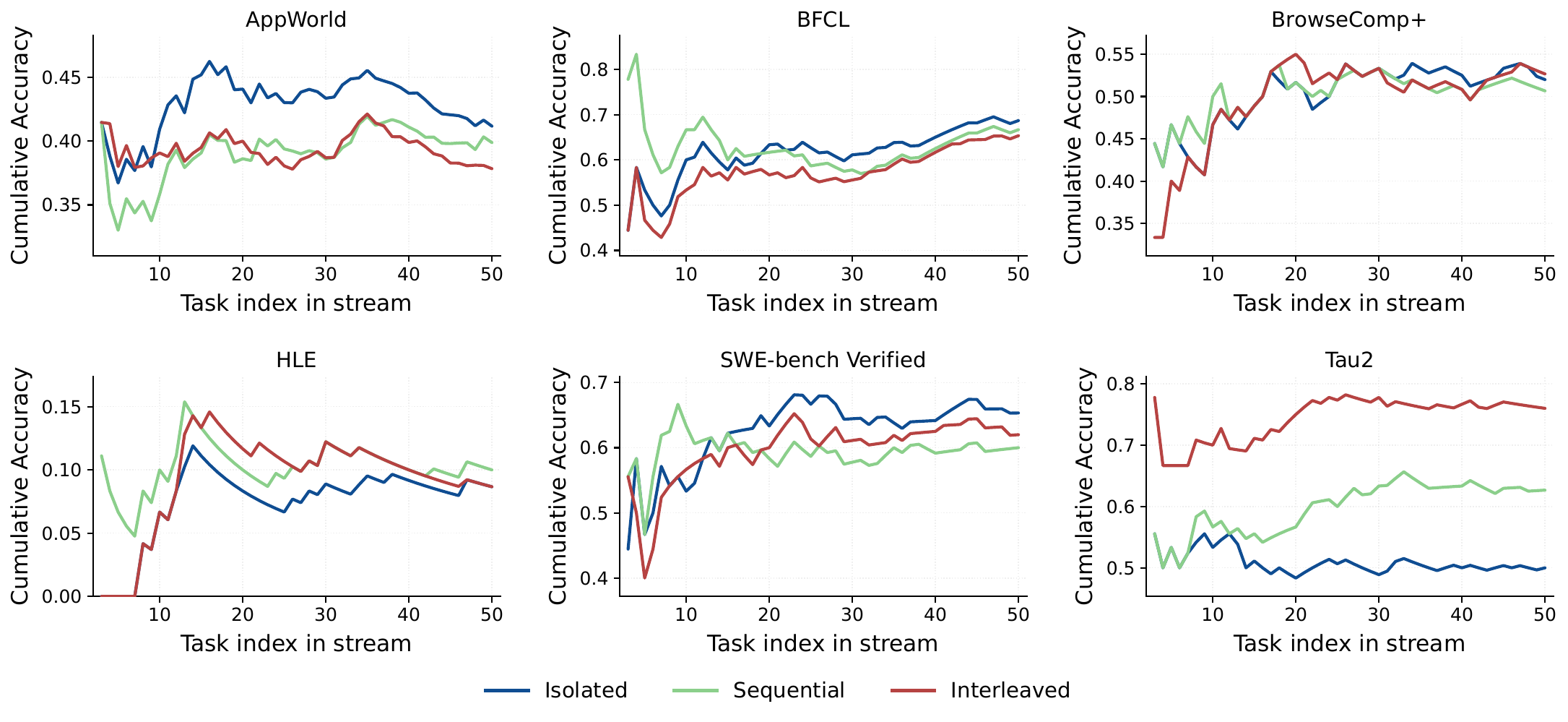}
    \caption{Cumulative accuracy dynamics of A-Mem on GPT-5.4 over three streaming scenarios.}
    \label{fig:cumulative_amem_gpt}
\end{figure}

\begin{figure}[!htbp]
    \centering
    \includegraphics[width=0.9\textwidth]{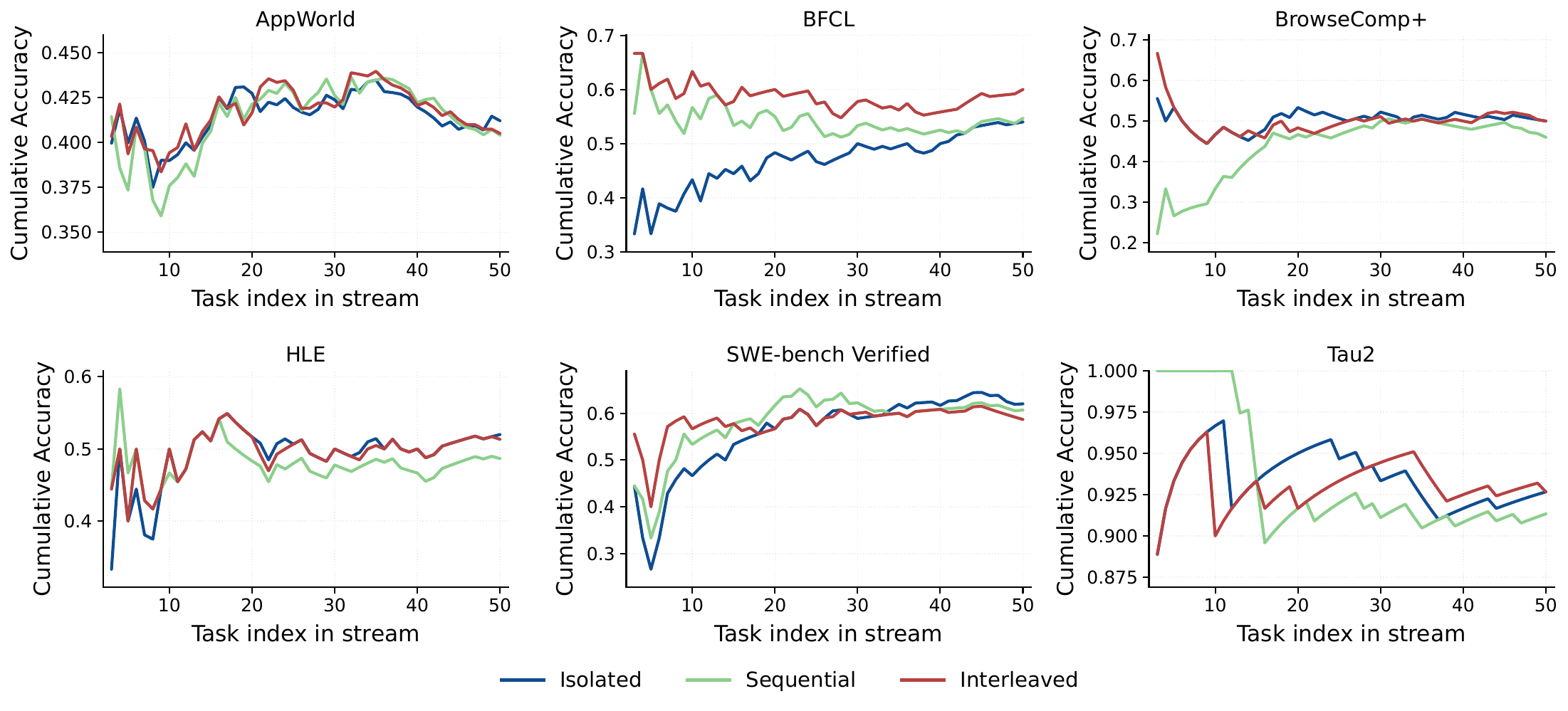}
    \caption{Cumulative accuracy dynamics of A-Mem on Gemini 3.1 Pro over three streaming scenarios.}
    \label{fig:cumulative_amem_gemini}
\end{figure}

\begin{figure}[!htbp]
    \centering
    \includegraphics[width=0.9\textwidth]{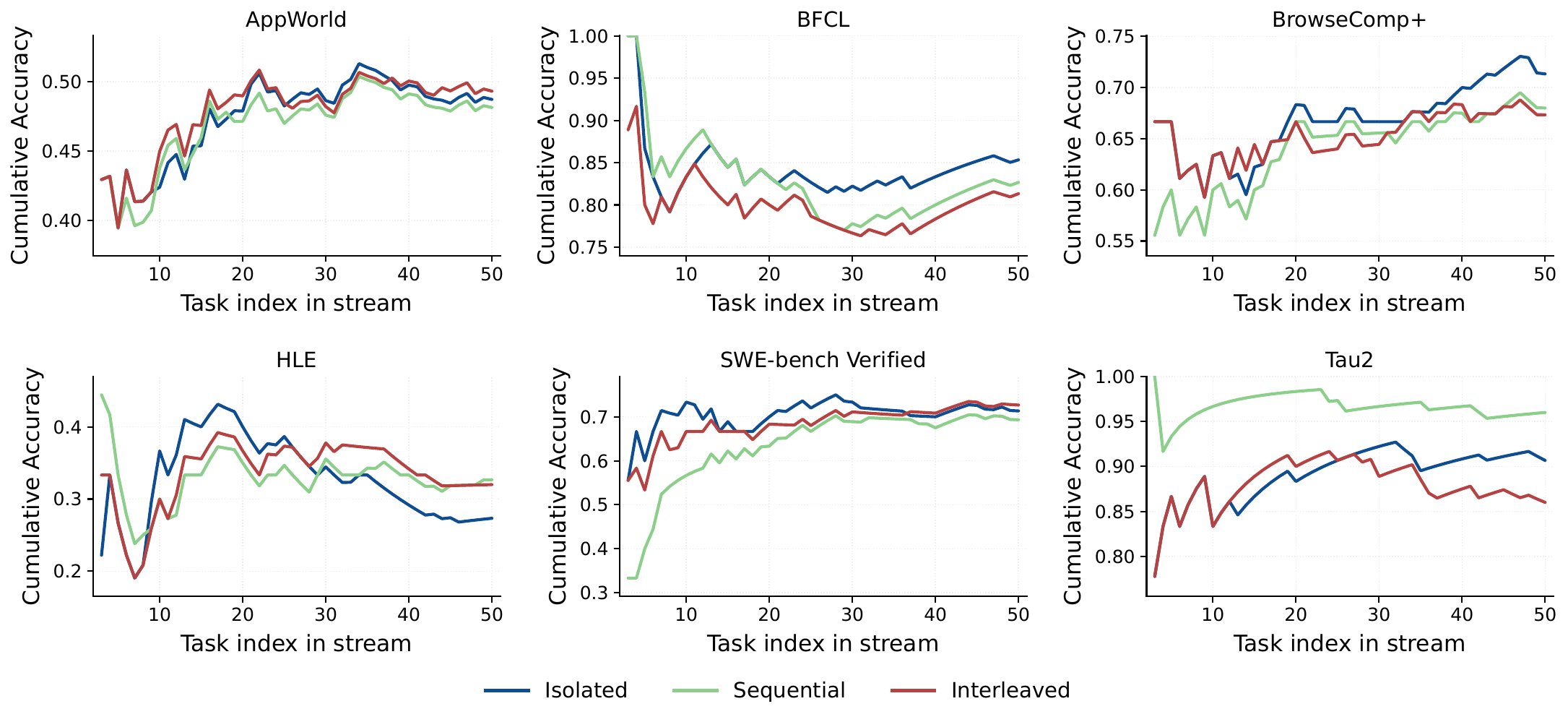}
    \caption{Cumulative accuracy dynamics of A-Mem on Claude Opus 4.7 over three streaming scenarios.}
    \label{fig:cumulative_amem_claude}
\end{figure}

\begin{figure}[!htbp]
    \centering
    \includegraphics[width=0.9\textwidth]{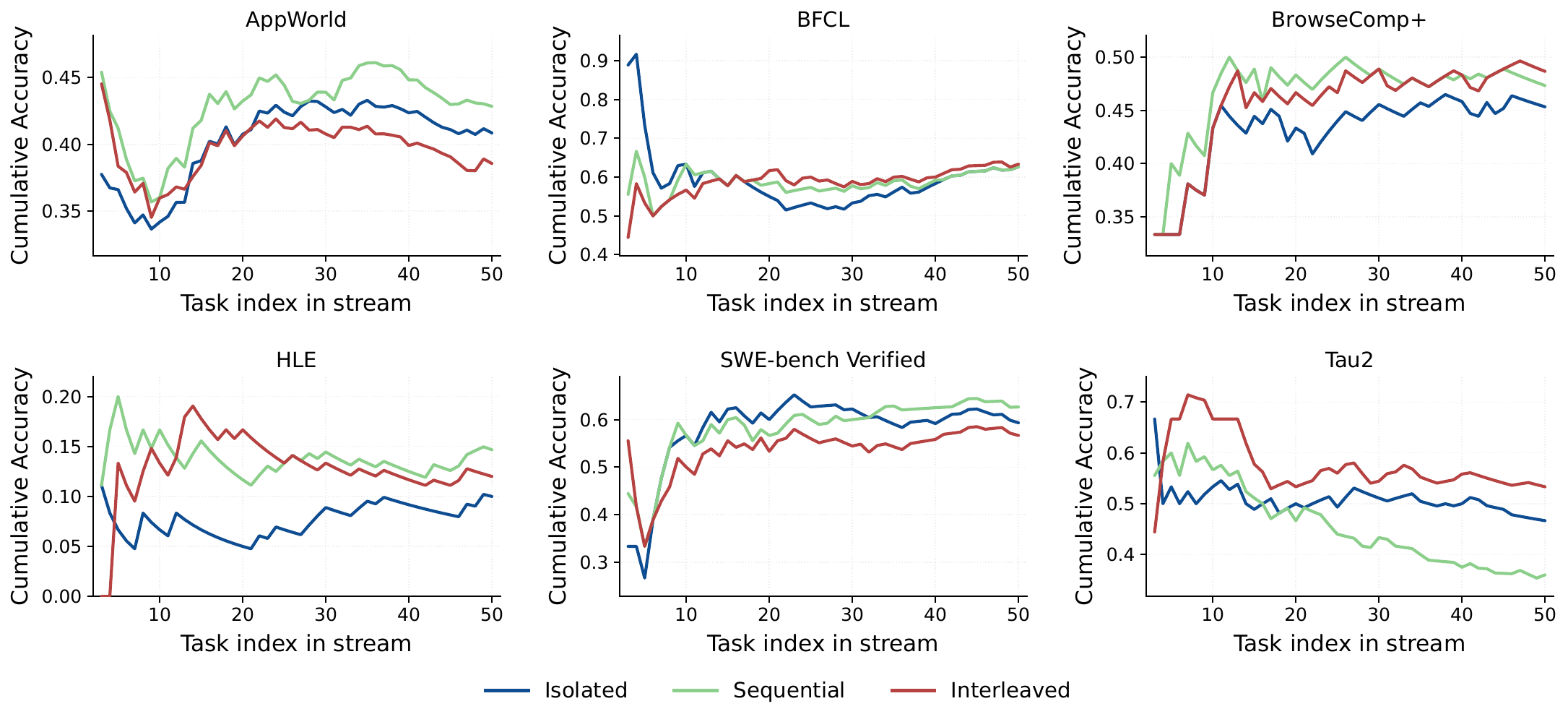}
    \caption{Cumulative accuracy dynamics of ReasoningBank on GPT-5.4 over three streaming scenarios.}
    \label{fig:cumulative_reasoningbank_gpt}
\end{figure}

\begin{figure}[!htbp]
    \centering
    \includegraphics[width=0.9\textwidth]{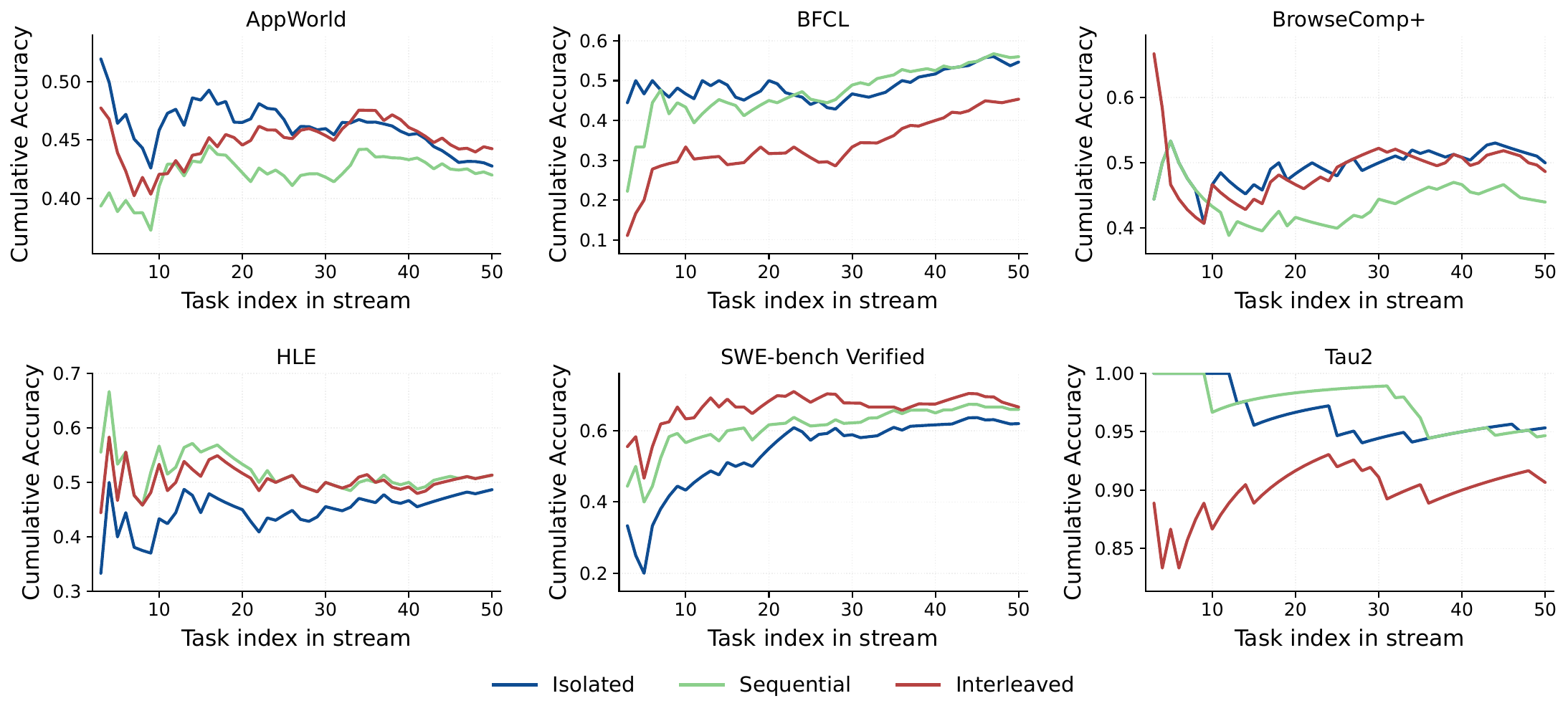}
    \caption{Cumulative accuracy dynamics of ReasoningBank on Gemini 3.1 Pro over three streaming scenarios.}
    \label{fig:cumulative_reasoningbank_gemini}
\end{figure}

\begin{figure}[!htbp]
    \centering
    \includegraphics[width=0.9\textwidth]{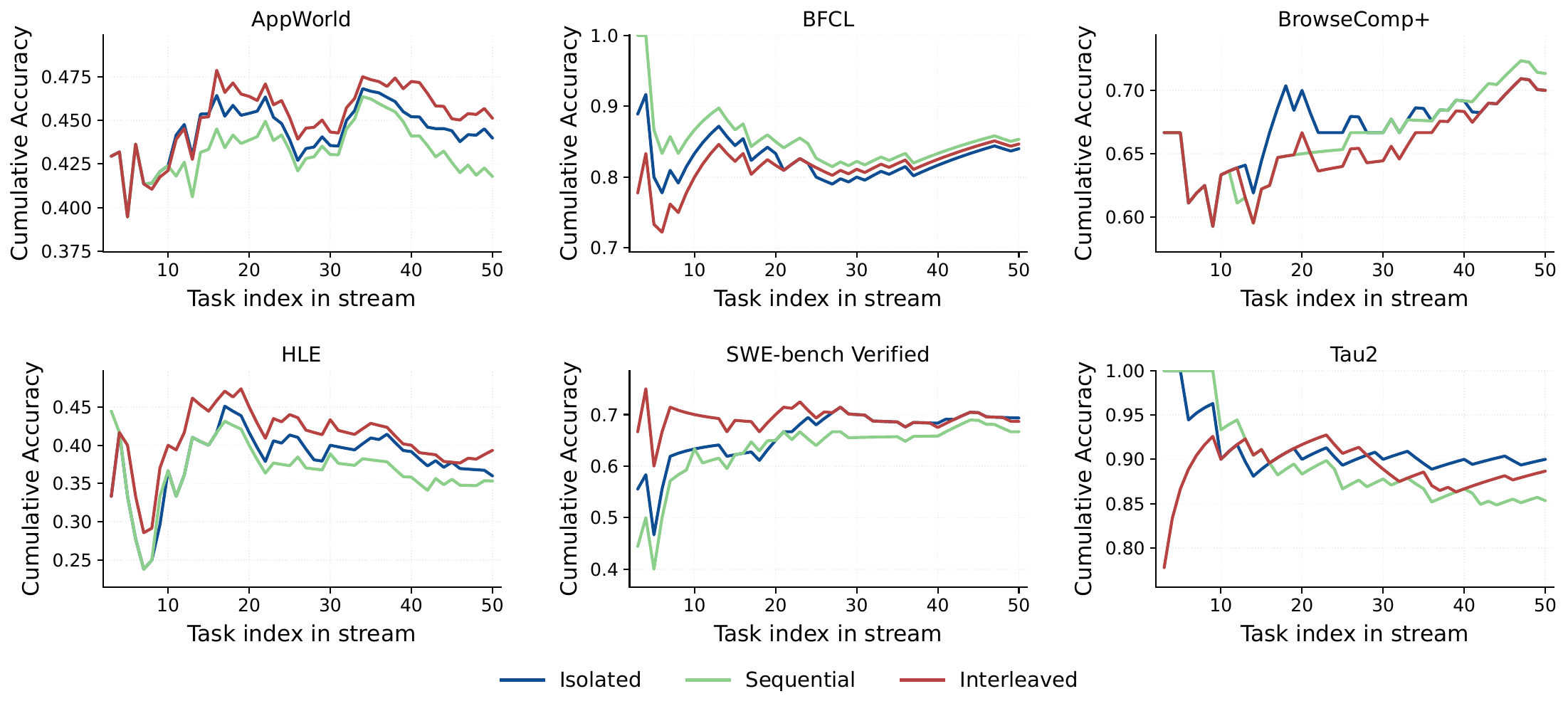}
    \caption{Cumulative accuracy dynamics of ReasoningBank on Claude Opus 4.7 over three streaming scenarios.}
    \label{fig:cumulative_reasoningbank_claude}
\end{figure}

\begin{figure}[!htbp]
    \centering
    \includegraphics[width=0.9\textwidth]{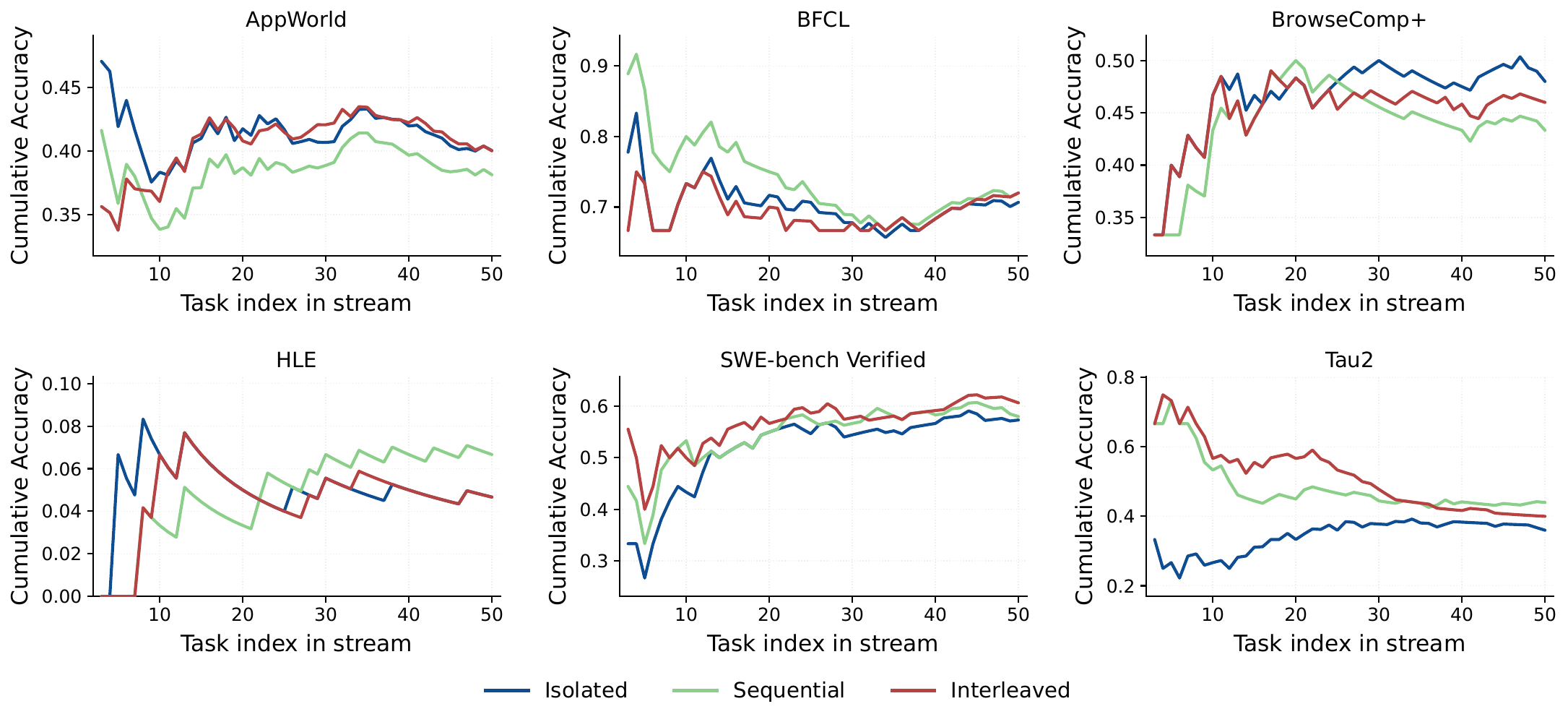}
    \caption{Cumulative accuracy dynamics of AutoSkill on GPT-5.4 over three streaming scenarios.}
    \label{fig:cumulative_autoskill_gpt}
\end{figure}

\begin{figure}[!htbp]
    \centering
    \includegraphics[width=0.9\textwidth]{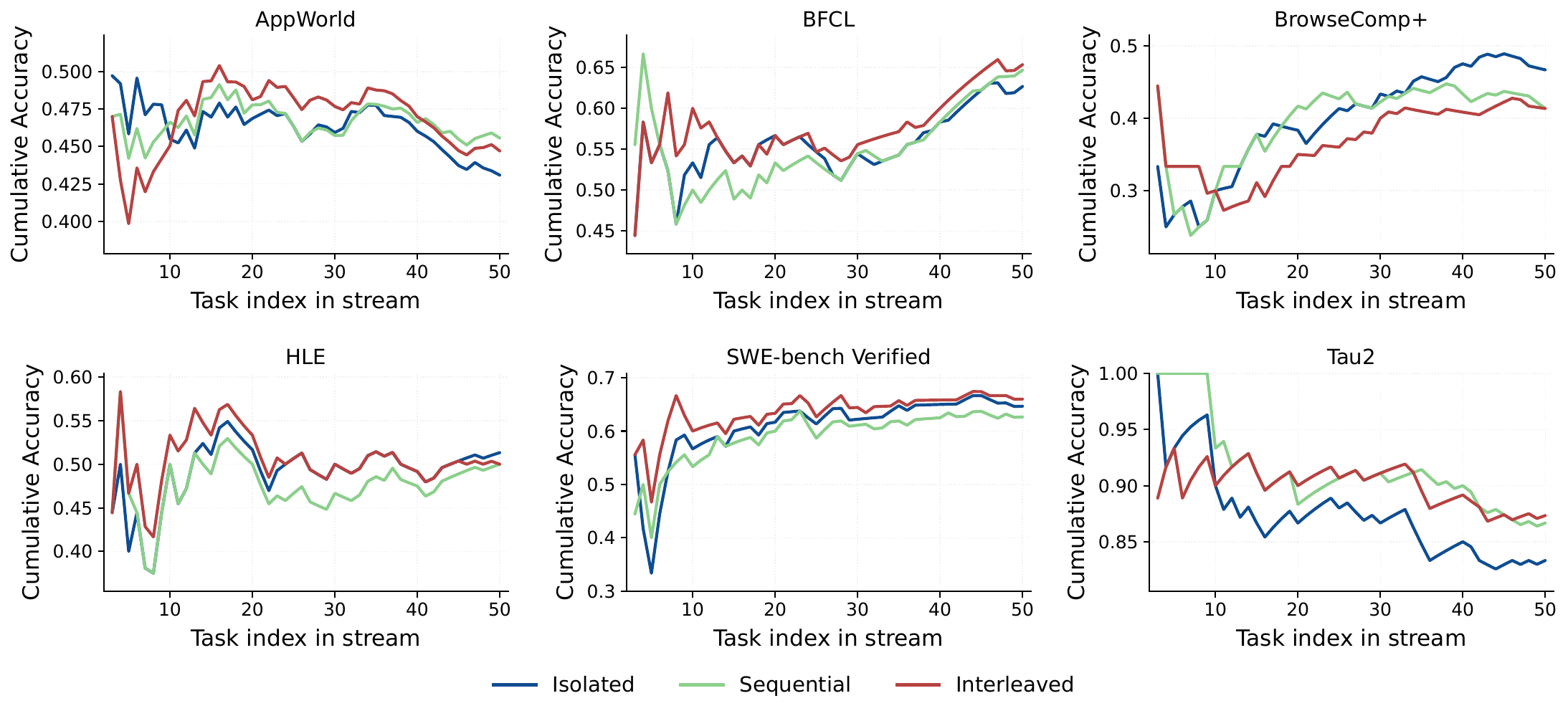}
    \caption{Cumulative accuracy dynamics of AutoSkill on Gemini 3.1 Pro over three streaming scenarios.}
    \label{fig:cumulative_autoskill_gemini}
\end{figure}

\begin{figure}[!htbp]
    \centering
    \includegraphics[width=0.9\textwidth]{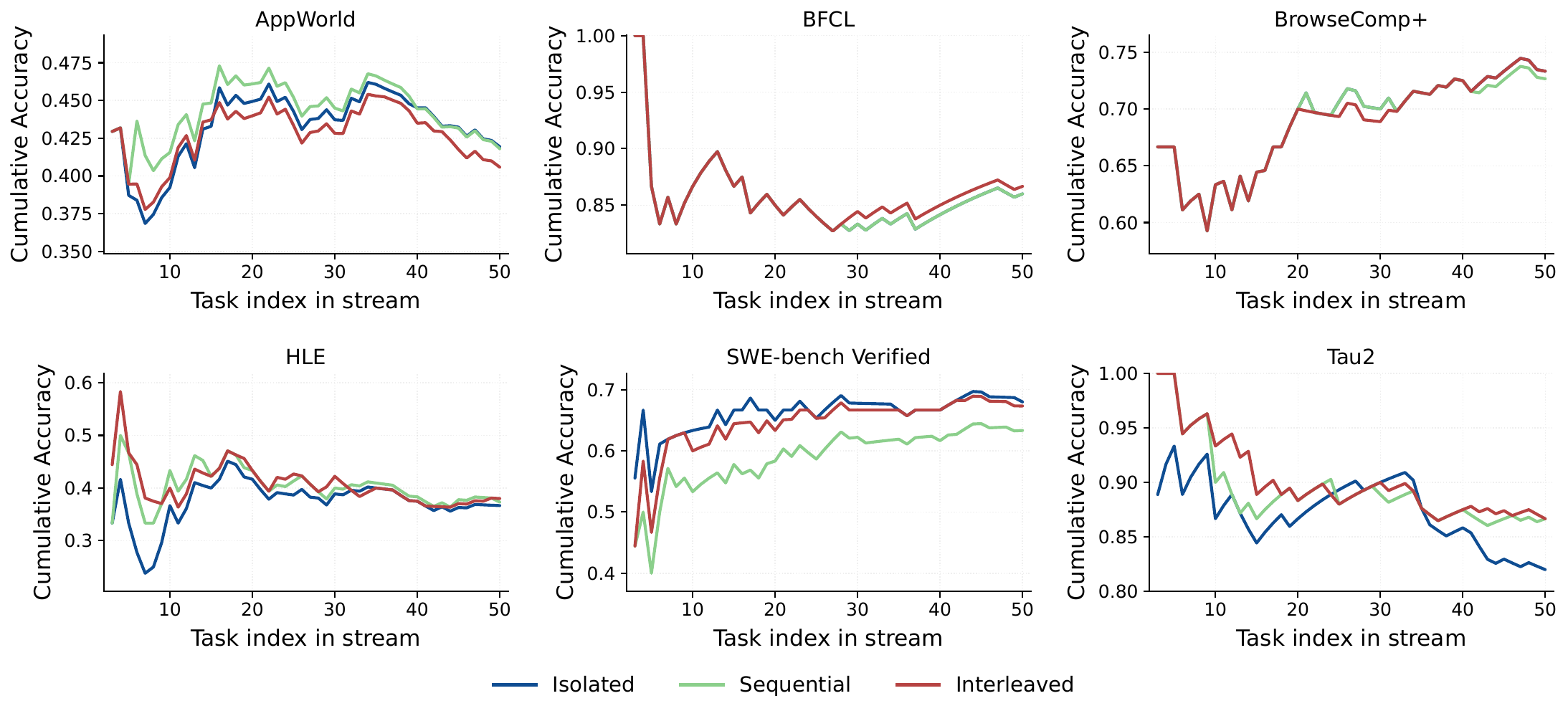}
    \caption{Cumulative accuracy dynamics of AutoSkill on Claude Opus 4.7 over three streaming scenarios.}
    \label{fig:cumulative_autoskill_claude}
\end{figure}

\begin{figure}[!htbp]
    \centering
    \includegraphics[width=0.9\textwidth]{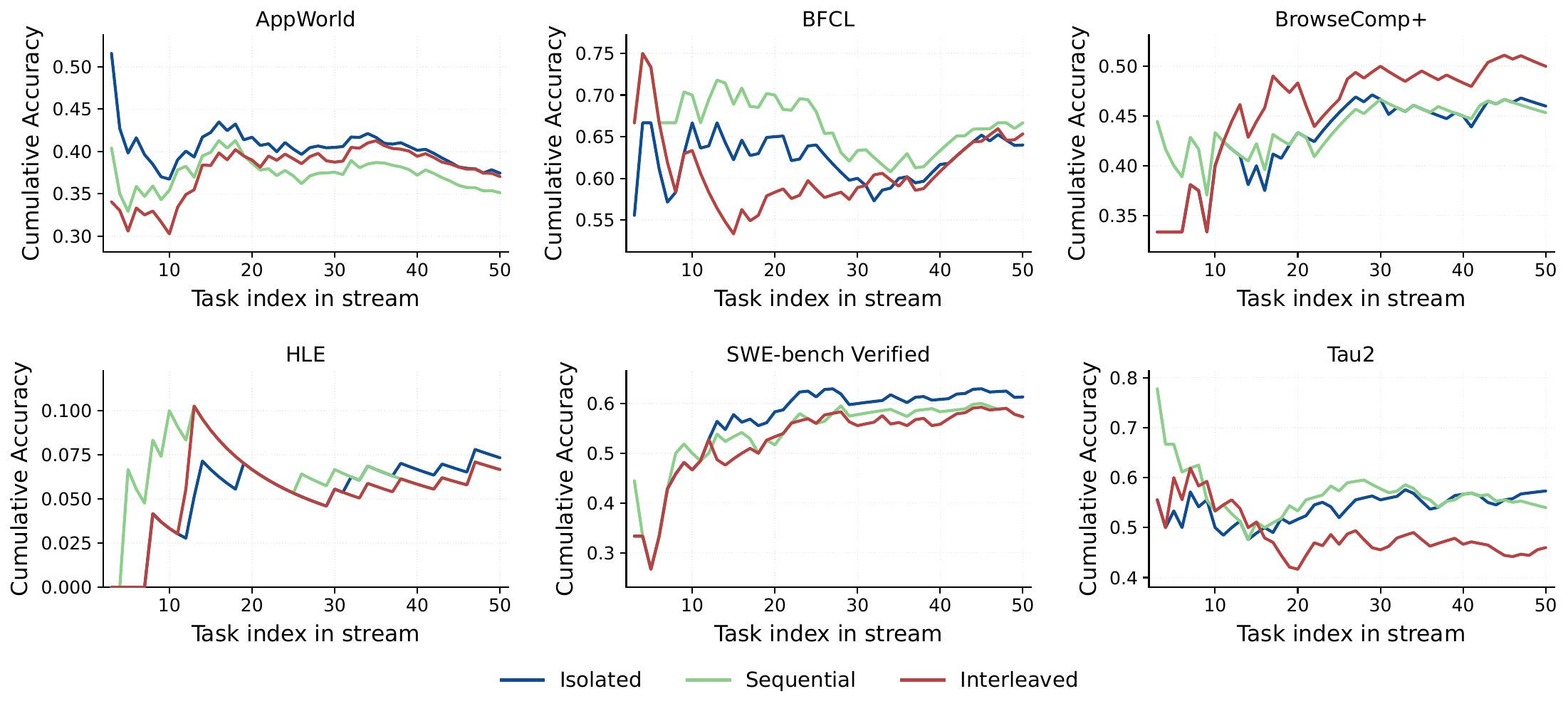}
    \caption{Cumulative accuracy dynamics of Harness on GPT-5.4 over three streaming scenarios.}
    \label{fig:cumulative_harness_gpt}
\end{figure}

\begin{figure}[!htbp]
    \centering
    \includegraphics[width=0.9\textwidth]{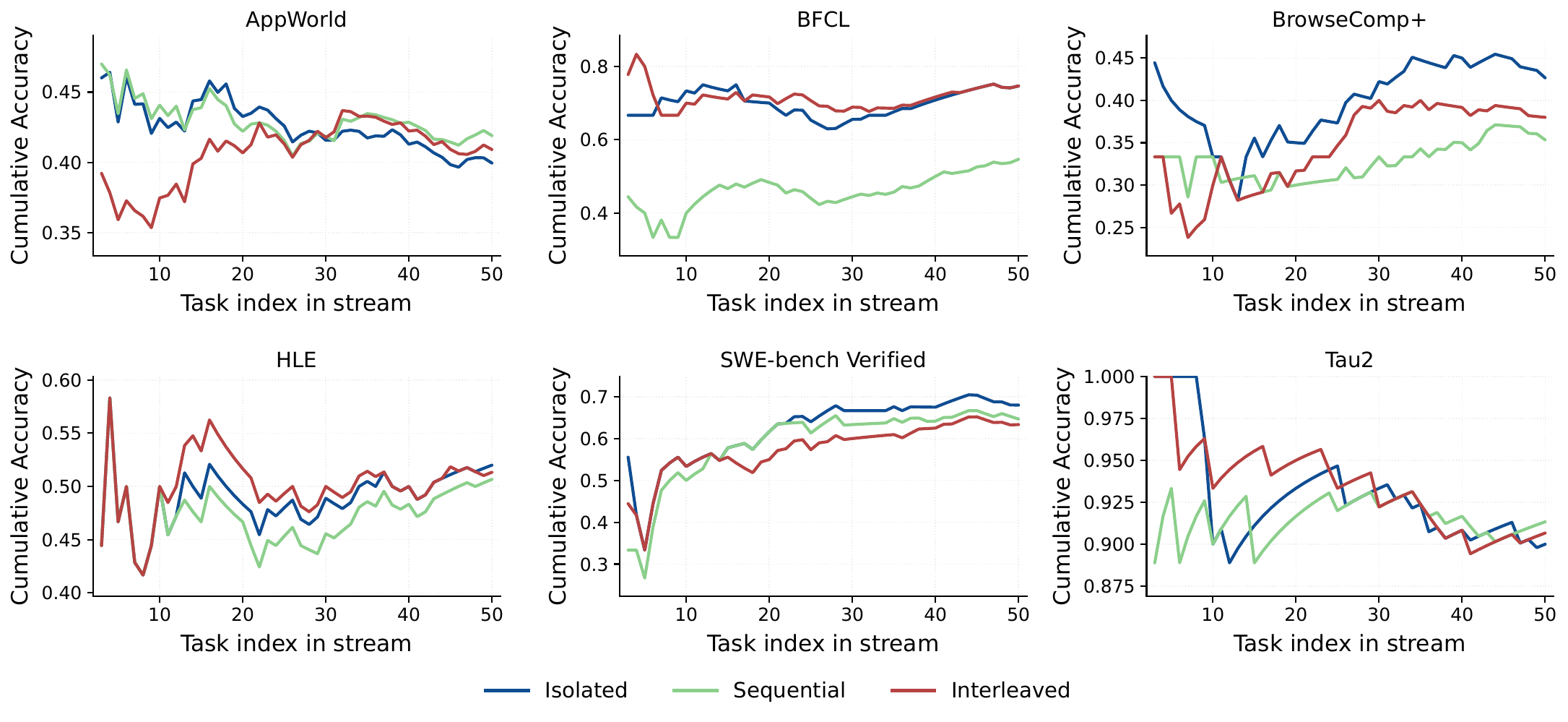}
    \caption{Cumulative accuracy dynamics of Harness on Gemini 3.1 Pro over three streaming scenarios.}
    \label{fig:cumulative_harness_gemini}
\end{figure}

\begin{figure}[!htbp]
    \centering
    \includegraphics[width=0.9\textwidth]{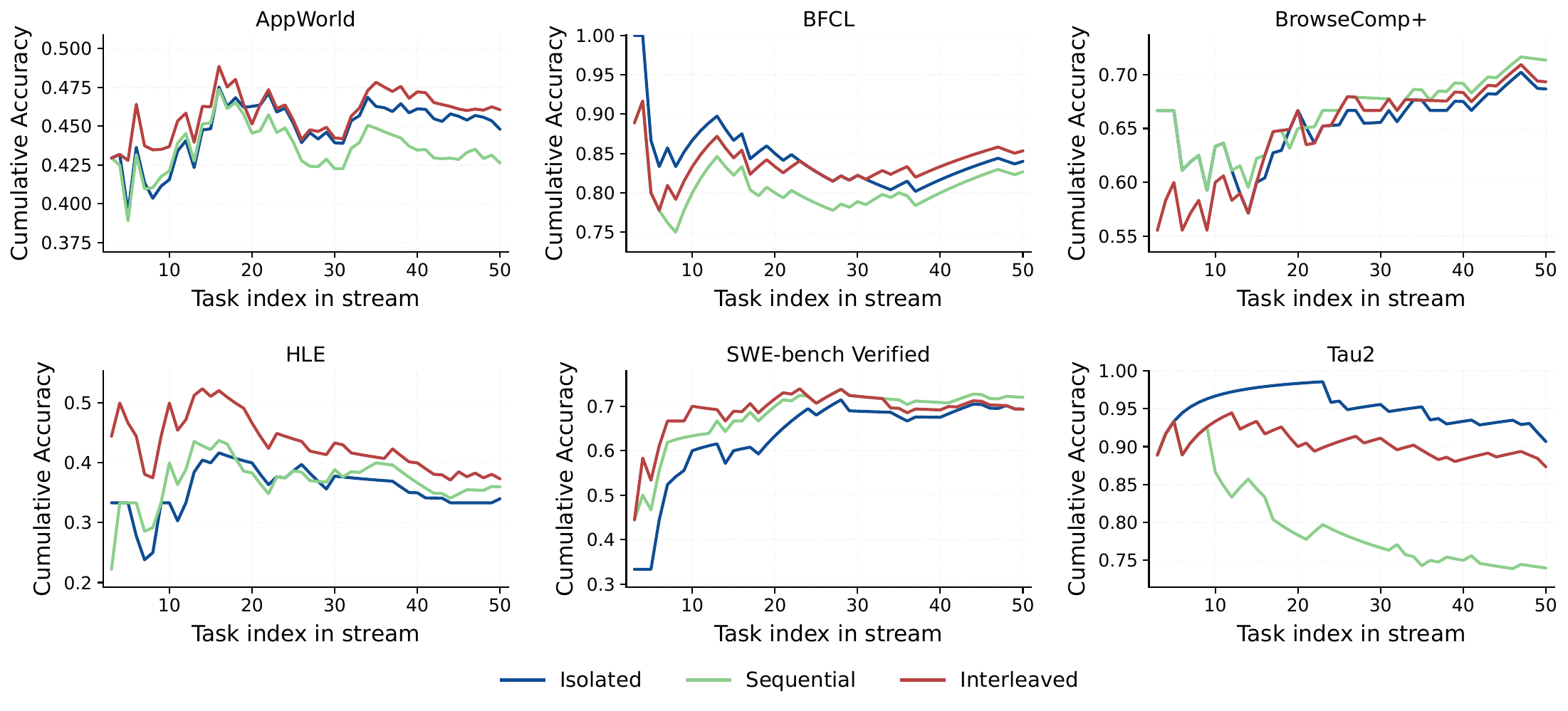}
    \caption{Cumulative accuracy dynamics of Harness on Claude Opus 4.7 over three streaming scenarios.}
    \label{fig:cumulative_harness_claude}
\end{figure}

To characterize the dynamics of self-evolution, we report the cumulative accuracy over three streaming scenarios for each self-evolving method and model in \crefrange{fig:cumulative_ace_gpt}{fig:cumulative_harness_claude}.

\section{Method Prompts}
\label{sec:method_prompts}
For reproducibility, we provide the full set of prompts used by self-evolving methods. All prompts are adapted to our test-time setting, where no ground-truth labels are available, and to the Exgentic~\citep{bandel2026general} framework, in which the agent completes tasks through multi-step tool use rather than single-turn question answering.

\subsection{ACE}
\label{sec:prompt_ace}

\begin{promptbox}[ACE Generator (used as the system prompt)]
\begin{verbatim}
You are an expert agent that completes tasks using available tools.
You have access to a curated playbook of strategies and insights learned from
previous tasks.  Use these to make better decisions.

## Guidelines
- Read the playbook carefully and apply relevant strategies
- Pay attention to common mistakes listed and avoid them
- Use available tools to interact with the environment
- Think step-by-step before acting
- When you are confident in your solution, use the finish/submit tool
- When a playbook bullet influences your decision, mention its ID
  (e.g. [err-00001]) in your reasoning text

## Playbook (accumulated strategies & insights)
Each line has a bullet ID and usage stats (helpful=N means it helped N times,
harmful=N means it misled N times).
Prefer high-helpful, low-harmful bullets.

{playbook}
\end{verbatim}
\end{promptbox}

\begin{promptbox}[ACE Reflector]
\begin{verbatim}
You are an expert analyst and educator. Your job is to analyze a model's
reasoning process and identify potential issues or strengths based on the
reasoning trace alone.

**Instructions:**
- Carefully analyze the model's reasoning trace to evaluate its approach
- The reasoning trace includes both the model's actions and environment
  observations in chronological order
- Identify potential conceptual errors, calculation mistakes, or misapplied
  strategies
- Also note what the model did well
- Provide actionable insights that could help the model perform better in
  future tasks
- Focus on the root cause, not just surface-level observations
- Be specific about what could be improved
- You will receive the full playbook that was available to the agent.
- Based on the reasoning trace, infer which bullets the agent likely applied
  or was influenced by, and tag each relevant bullet as 'helpful', 'harmful',
  or 'neutral'. Skip unrelated bullets.

Your output should be a json object, which contains the following fields
- reasoning: your chain of thought / reasoning / thinking process
- error_identification: what potential issues exist in the reasoning?
  (or "none identified" if the approach appears sound)
- root_cause_analysis: why might these issues occur? What concept may have
  been misunderstood?
- correct_approach: what could the model do differently or better?
- key_insight: what strategy, formula, or principle should be remembered for
  future tasks?
- bullet_tags: a list of json objects with bullet id and tag for each
  relevant playbook bullet

**Question:**
{question}

**Model's Reasoning Trace:**
{reasoning_trace}

**Model's Predicted Answer:**
{predicted_answer}

**Full Playbook:**
{bullets_used}

**Answer in this exact JSON format:**
{
  "reasoning": "[Your chain of thought / reasoning / thinking process]",
  "error_identification": "[What potential issues exist in the reasoning?]",
  "root_cause_analysis": "[Why might these issues occur?]",
  "correct_approach": "[What could the model do differently or better?]",
  "key_insight": "[What strategy or principle should be remembered?]",
  "bullet_tags": [
    {"id": "calc-00001", "tag": "helpful"},
    {"id": "fin-00002", "tag": "harmful"}
  ]
}
\end{verbatim}
\end{promptbox}

\begin{promptbox}[ACE Curator]
\begin{verbatim}
You are a master curator of knowledge. Your job is to identify what new
insights should be added to an existing playbook based on a reflection from
a previous attempt.

**Context:**
- The playbook you created will be used to help answering similar questions.
- The reflection is generated using environment feedback that will NOT be
  available when the playbook is being used.

**CRITICAL: You MUST respond with valid JSON only. Do not use markdown
formatting or code blocks.**

**Instructions:**
- Review the existing playbook and the reflection from the previous attempt
- Identify ONLY the NEW insights, strategies, or mistakes that are MISSING
  from the current playbook
- Avoid redundancy - if similar advice already exists, only add new content
  that is a perfect complement to the existing playbook
- Do NOT regenerate the entire playbook - only provide the additions needed
- Focus on quality over quantity - a focused, well-organized playbook is
  better than an exhaustive one
- Format your response as a PURE JSON object with specific sections
- For any operation if no new content to add, return an empty list for the
  operations field
- Be concise and specific - each addition should be actionable

**Training Context:**
- Total token budget: {token_budget} tokens
- Training progress: Sample {current_step} out of {total_samples}

**Current Playbook Stats:**
{playbook_stats}

**Recent Reflection:**
{recent_reflection}

**Current Playbook:**
{current_playbook}

**Question Context:**
{question_context}

**Your Task:**
Output ONLY a valid JSON object with these exact fields:
- reasoning: your chain of thought / reasoning / thinking process
- operations: a list of operations to be performed on the playbook
  - type: the type of operation to be performed
  - section: the section to add the bullet to
  - content: the new content of the bullet

**Available Operations:**
1. ADD: Create new bullet points with fresh IDs
    - section: the section to add the new bullet to
    - content: the new content of the bullet. Note: no need to include the
      bullet_id in the content like '[ctx-00263] helpful=1 harmful=0 ::',
      the bullet_id will be added by the system.

**RESPONSE FORMAT - Output ONLY this JSON structure (no markdown, no code
blocks):**
{
  "reasoning": "[Your reasoning here]",
  "operations": [
    {
      "type": "ADD",
      "section": "formulas_and_calculations",
      "content": "[New calculation method...]"
    }
  ]
}
\end{verbatim}
\end{promptbox}

\subsection{A-Mem}
\label{sec:prompt_amem}

\begin{promptbox}[A-Mem Generator (used as the system prompt)]
\begin{verbatim}
You are an expert agent that completes tasks using available tools.
Think step-by-step before acting.
Use available tools to interact with the environment.
When you are confident in your solution, use the finish/submit tool.

Based on the context below, complete the task. Use the context to inform
your decisions.

Context:
{memory_context}
\end{verbatim}
\end{promptbox}

\begin{promptbox}[A-Mem Retrieval Query]
\begin{verbatim}
Given the following question, generate several keywords separated by commas.

Question: {question}

Keywords:
\end{verbatim}
\end{promptbox}

\begin{promptbox}[A-Mem Content Analysis]
\begin{verbatim}
Analyze the following content and provide:
1. KEYWORDS: The most important keywords (nouns, verbs, key concepts). Order
   from most to least important. At least three keywords. Do not include
   speaker names or time references.
2. CONTEXT: One sentence summarizing the main topic, key points, and purpose.
3. TAGS: Broad categories/themes for classification (domain, format, type).
   At least three tags.

Respond using EXACTLY this format (one section per header):

KEYWORDS: keyword1, keyword2, keyword3, ...
CONTEXT: A single sentence summarizing the content.
TAGS: tag1, tag2, tag3, ...

Content for analysis:
{content}
\end{verbatim}
\end{promptbox}

\begin{promptbox}[A-Mem Focused Keywords]
\begin{verbatim}
List exactly 5 keywords that capture the main concepts of the following text.
Output only the keywords, comma-separated, nothing else.

Text: {content}
\end{verbatim}
\end{promptbox}

\begin{promptbox}[A-Mem Evolution Decision]
\begin{verbatim}
You are an AI memory evolution agent. Analyze the new memory note and its
nearest neighbors to decide if evolution is needed.

New memory:
- Context: {context}
- Content: {content}
- Keywords: {keywords}

Nearest neighbor memories:
{nearest_neighbors_memories}

Based on the relationships between the new memory and its neighbors, decide:
- NO_EVOLUTION: The memory stands alone, no changes needed.
- STRENGTHEN: The new memory should be linked to some neighbors and its tags
  updated.
- UPDATE_NEIGHBOR: The neighbors' context/tags should be updated based on new
  understanding.
- STRENGTHEN_AND_UPDATE: Both strengthen and update neighbors.

Respond using EXACTLY this format:
DECISION: <one of NO_EVOLUTION, STRENGTHEN, UPDATE_NEIGHBOR,
STRENGTHEN_AND_UPDATE>
REASON: <brief explanation>
\end{verbatim}
\end{promptbox}

\begin{promptbox}[A-Mem Strengthen Details]
\begin{verbatim}
Given the new memory and its neighbors, provide updated connections and tags.

New memory:
- Content: {content}
- Keywords: {keywords}

Neighbor memories:
{nearest_neighbors_memories}

Which neighbor indices should the new memory connect to? What tags best
describe this memory?

Respond using EXACTLY this format:
CONNECTIONS: 0, 2, 3
TAGS: tag1, tag2, tag3, ...
\end{verbatim}
\end{promptbox}

\begin{promptbox}[A-Mem Update Neighbors]
\begin{verbatim}
Given the new memory and its neighbor memories, update each neighbor's
context and tags based on a holistic understanding of all these memories
together.

New memory:
- Content: {content}
- Context: {context}

Neighbor memories:
{nearest_neighbors_memories}

For each neighbor (indexed 0 to {max_neighbor_idx}), provide updated context
and tags. If no change is needed, repeat the original values.

Respond using EXACTLY this format (one block per neighbor):

NEIGHBOR 0:
CONTEXT: updated context sentence
TAGS: tag1, tag2, tag3

NEIGHBOR 1:
CONTEXT: updated context sentence
TAGS: tag1, tag2, tag3

(continue for all {neighbor_count} neighbors)
\end{verbatim}
\end{promptbox}

\subsection{ReasoningBank}
\label{sec:prompt_reasoningbank}

\begin{promptbox}[ReasoningBank Generator (used as the system prompt)]
\begin{verbatim}
You are an expert agent that completes tasks using available tools.

Below are some memory items that I accumulated from past interaction from the
environment that may be helpful to solve the task. You can use it when you 
feel it's relevant. In each step, please first explicitly discuss if you want
to use each memory item or not, and then take action.

{retrieved_memory_items}
\end{verbatim}
\end{promptbox}

\begin{promptbox}[ReasoningBank Trajectory Evaluator (system message)]
\begin{verbatim}
You are an expert in evaluating the performance of a task-solving agent. The
agent is designed to help a human user complete a task by taking actions in 
an environment. Given the user's intent, the agent's action history, the
environment's feedback, and the agent's response to the user, your goal is to
decide whether the agent's execution is successful or not.

*Strictness rules*
Before calling a task successful, verify all three:
- Completeness: every constraint in the intent is satisfied.
- Grounding: every value or result the agent reports is traceable to a 
  specific observation from the environment; values that were inferred,
  guessed, or summarized without a visible source count as failures.
- Right target: when the task names a specific entity, confirm the agent
  acted on that exact entity and not an adjacent one.
When uncertain on any of these, mark failure. A false success is more harmful
than a false failure, because memory induction amplifies it into future
behavior.

*IMPORTANT*
Format your response into two lines as shown below:

Thoughts: <your thoughts and reasoning process>"
Status: "success" or "failure"
\end{verbatim}
\end{promptbox}

\begin{promptbox}[ReasoningBank Trajectory Evaluator (user message)]
\begin{verbatim}
User Intent: {intent}

Action History:
{last_actions}

Environment feedback (last observations):

```
{cap}
```

Agent response to the user: {response}.
\end{verbatim}
\end{promptbox}

\begin{promptbox}[ReasoningBank Memory Induction, success (system message)]
\begin{verbatim}
You are an expert at analyzing agent task execution. You will be given a user
query, the corresponding trajectory that represents **how an agent 
successfully accomplished the task**.

## Guidelines
You need to extract and summarize useful insights in the format of memory 
items based on the agent's successful trajectory.
The goal of summarized memory items is to be helpful and generalizable for
future similar tasks.

## Important notes
  - You must first think why the trajectory is successful, and then summarize
    the insights.
  - You can extract *at most 3* memory items from the trajectory.
  - You must not repeat similar or overlapping items.
  - Prefer concrete, actionable procedures over abstract principles. Do not
    embed specific product names, queries, or literal string contents from 
    the task.

## Output Format
Your output must strictly follow the Markdown format shown below:

# Memory Item i
## Title <the title of the memory item>
## Description <one sentence summary describing when or when NOT to use the
memory item>
## Content <1-3 sentences describing the insights learned to successfully
accomplishing similar tasks in the future>
\end{verbatim}
\end{promptbox}

\begin{promptbox}[ReasoningBank Memory Induction, failure (system message)]
\begin{verbatim}
You are an expert at analyzing agent task execution. You will be given a user
query, the corresponding trajectory that represents **how an agent attempted 
to resolve the task but failed**.

## Guidelines
You need to extract and summarize useful insights in the format of memory 
items based on the agent's failed trajectory.
The goal of summarized memory items is to be helpful and generalizable for
future similar tasks.

## Important notes
  - You must first reflect and think why the trajectory failed, and then
    summarize what lessons you have learned or strategies to prevent the
    failure in the future.
  - You can extract *at most 3* memory items from the trajectory.
  - You must not repeat similar or overlapping items.
  - Prefer concrete, actionable recovery procedures over abstract principles.
    Do not embed specific product names, queries, or literal string contents
    from the task.

## Output Format
Your output must strictly follow the Markdown format shown below:

# Memory Item i
## Title <the title of the memory item>
## Description <one sentence summary describing when or when NOT to use the
memory item>
## Content <1-3 sentences describing the insights learned to avoid such
failures and successfully accomplishing similar tasks in the future>
\end{verbatim}
\end{promptbox}

\begin{promptbox}[ReasoningBank Memory Induction (user message)]
\begin{verbatim}
**Query:** {query}

**Trajectory:**
{trajectory}
\end{verbatim}
\end{promptbox}

\subsection{AutoSkill}
\label{sec:prompt_autoskill}

\begin{promptbox}[AutoSkill Generator (used as the system prompt)]
\begin{verbatim}
You are an expert agent that completes tasks using available tools.
Think step-by-step before acting.
Use available tools to interact with the environment.
When you are confident in your solution, use the finish/submit tool.

## Retrieved Skills (from accumulated experience)
The following skills were retrieved based on relevance to the current task.
Use a skill ONLY when it directly matches the current intent. Otherwise, 
ignore all retrieved skills and act normally. Never explicitly mention that 
skills were retrieved/injected.

{skills_block}
\end{verbatim}
\end{promptbox}

\begin{promptbox}[AutoSkill Retrieved Skill Entry]
\begin{verbatim}
### Skill: {name}
- **Description**: {description}
- **Tags**: {tags}
- **Triggers**: {triggers}

**Instructions**:
{instructions}
\end{verbatim}
\end{promptbox}

\begin{promptbox}[AutoSkill Query Rewrite]
\begin{verbatim}
You are a retrieval query rewriter. Your job is to rewrite the current user 
task into a concise, standalone search query for skill retrieval.

Core rules:
- Produce exactly ONE line of output: the rewritten query.
- Resolve references ("it", "this", "the above") using the provided context.
- Keep only retrieval-relevant constraints (format, audience, quality,
  domain).
- Preserve the task anchor (what the task is about).
- Do NOT include generic process words without a concrete topic anchor.

Task: {task}
Context: {context}

Rewritten query:
\end{verbatim}
\end{promptbox}

\begin{promptbox}[AutoSkill Skill Extraction]
\begin{verbatim}
You are a skill extractor that turns agent interaction traces into reusable
skills.

## Extraction Principles
- Treat the task description and environment observations as primary evidence.
- Extract ONLY when there are durable, reusable constraints, policies,
  workflows, or strategies that would help in FUTURE similar tasks.
- Do NOT extract one-shot task-specific facts or generic "be helpful" 
  patterns.
- Capture HOW TO DO similar tasks, rather than this-instance facts.
- Remove case-specific entities (names, URLs, dates) and preserve only 
  portable rules.
- Do NOT invent workflow steps unless explicitly demonstrated in the trace.
- If nothing reusable is found, return an empty skills list.

## Session Information
Task: {task}

## Session Trace (Actions & Observations)
{session_trace}

## Output Format
Return a JSON object with this schema:
{
  "skills": [
    {
      "name": "<concise, searchable name>",
      "description": "<what this skill does and when to use it>",
      "instructions": "<markdown body with # Goal, # Constraints & Style,
       # Workflow (optional)>",
      "triggers": ["<intent phrase 1>", "<intent phrase 2>", ...],
      "tags": ["<keyword1>", "<keyword2>", ...],
      "confidence": <float 0.0-1.0>
    }
  ]
}

If nothing reusable is detected, return: {"skills": []}
\end{verbatim}
\end{promptbox}

\begin{promptbox}[AutoSkill Skill Judge]
\begin{verbatim}
You are a skill set manager. Given a newly extracted skill candidate and 
the most similar existing skill from the skill bank, decide the appropriate
action.

## Decision Procedure
1. Check if the candidate represents the same capability as the existing skill
   (same job-to-be-done, same deliverable type, overlapping constraints).
2. Apply discard gate: reject generic, low-signal, non-portable candidates.
3. Compare on four axes: job-to-be-done, deliverable type, hard
   constraints/success criteria, and required tools/workflow.
4. Choose "merge" ONLY when they are the same capability after removing
   instance details.
5. Choose "add" when the candidate is a distinct durable capability.
6. Choose "discard" when the candidate is too generic or non-reusable.

## Candidate Skill
Name: {candidate_name}
Description: {candidate_description}
Instructions: {candidate_instructions}
Triggers: {candidate_triggers}
Tags: {candidate_tags}

## Most Similar Existing Skill (may be empty if no skills exist)
Name: {existing_name}
Description: {existing_description}
Instructions: {existing_instructions}
Triggers: {existing_triggers}
Tags: {existing_tags}
Similarity Score: {similarity_score}

## Output Format
Return a JSON object:
{
  "action": "add" | "merge" | "discard",
  "target_skill_id": "<id of existing skill to merge with, or null>",
  "reason": "<brief explanation>"
}
\end{verbatim}
\end{promptbox}

\begin{promptbox}[AutoSkill Skill Merge]
\begin{verbatim}
You are a skill merger. Combine an existing skill with a new candidate into
one improved skill that preserves the best of both.

## Merge Rules
- Preserve the original capability identity (name and core goal).
- Perform semantic union rather than raw concatenation.
- Import only reusable, non-conflicting additions from the candidate.
- Avoid regressions: keep important checks from the existing skill.
- Remove case-specific entities and one-off facts.
- Do NOT invent any new standards or details not present in either skill.
- Deduplicate sections, bullets, triggers, tags.
- Keep language consistent across all fields.

## Existing Skill
Name: {existing_name}
Description: {existing_description}
Instructions: {existing_instructions}
Triggers: {existing_triggers}
Tags: {existing_tags}

## Candidate Skill (new evidence)
Name: {candidate_name}
Description: {candidate_description}
Instructions: {candidate_instructions}
Triggers: {candidate_triggers}
Tags: {candidate_tags}

## Output Format
Return a JSON object with the merged skill:
{
  "name": "<merged name>",
  "description": "<merged description>",
  "instructions": "<merged instructions (markdown)>",
  "triggers": ["<trigger1>", ...],
  "tags": ["<tag1>", ...]
}
\end{verbatim}
\end{promptbox}

\subsection{Harness}
\label{sec:prompt_harness}

\begin{promptbox}[Harness Generator (used as the system prompt)]
\begin{verbatim}
You are an expert agent that completes tasks using available tools.
Think step-by-step before acting.
Use available tools to interact with the environment.
When you are confident in your solution, use the finish/submit tool.

## Long-Term Memory
{memory}

## Retrieved Skills
### Skill: {skill_name}
*{skill_description}*

{skill_body}
\end{verbatim}
\end{promptbox}

\begin{promptbox}[Harness Evolver (system prompt)]
\begin{verbatim}
You are an evolution engine for an agent harness. Your job is to analyze a
completed task session and improve the agent's harness (system prompt,
long-term memory, and skill library) for future tasks.

## Available Tools

**Read tools** (use these first to inspect current state):
- read_prompt() - read the current system prompt
- read_memory() - read the current long-term memory document
- list_skills() - list all skills with names and descriptions
- read_skill(name) - read a specific skill's full body

**Write tools** (use these to make changes):
- edit_prompt(body) - replace the entire system prompt
- edit_memory(body) - replace the entire memory document
- add_skill(name, description, body) - add a new skill
- edit_skill(name, description, body) - modify an existing skill
- delete_skill(name) - remove a skill

## Constraints
- At most 1 edit_prompt call per session.
- At most 1 edit_memory call per session.
- No limit on skill operations.

## Guidelines
- First READ the current harness state, then decide what changes to make.
- Skills should be generalizable (useful across tasks), not task-specific.
- Memory should capture recurring patterns, proven strategies, and environment
  quirks.
- System prompt changes should refine the agent's general approach.
- Do NOT duplicate information already present in the harness.
- If no changes are needed, simply stop without calling any write tools.
\end{verbatim}
\end{promptbox}

\begin{promptbox}[Harness Evolver (user message)]
\begin{verbatim}
## This Session

### Task
{task}

### Skills Injected
{injected_skill_names}

### Session Trajectory
{trajectory}

---

Analyze the session above. Read the current harness state using the read 
tools,then decide what changes (if any) would improve the agent's future 
performance. Make changes using the write tools, or stop if no changes 
are needed.
\end{verbatim}
\end{promptbox}

\end{document}